\documentclass{article}

\PassOptionsToPackage{numbers, compress}{natbib}

\usepackage[eandd,final]{neurips_2026}

\usepackage[utf8]{inputenc} 
\usepackage[T1]{fontenc}    
\usepackage{hyperref}       
\usepackage{url}            
\usepackage{booktabs}       
\usepackage{amsfonts}       
\usepackage{nicefrac}       
\usepackage{microtype}      
\usepackage{xcolor}         
\usepackage{multirow}
\usepackage{makecell}
\usepackage{longtable}
\usepackage{geometry}
\usepackage{graphicx}
\usepackage{wrapfig}

\usepackage{amsmath}
\usepackage{amssymb}
\usepackage{algorithm}
\usepackage{algpseudocode}

\usepackage{threeparttable}
\usepackage{subcaption}
\title{DeepfakeGenome: Toward Next-Generation \\ Deepfake Attribution}

\author{
Shunli Wang$^1$\thanks{Equal contribution.} ,
~ Xinyu Zhou$^{2~\ast}$,
~ Yandan Zhao$^1$,
~ Wenbin Liang$^{2,3}$,
~ Taiping Yao$^1$ \\
~\textbf{ Ke-Yue Zhang}$^1$\textbf{,}
~ \textbf{Zhiyuan Yan}$^4$\textbf{,}
~ \textbf{Shouhong Ding}$^1$\thanks{Corresponding author.}\textbf{ ,}
~ \textbf{Lizhuang Ma}$^{2~\dagger}$ \\
\normalsize $^1$ Tencent Youtu Lab \quad
$^2$ Shanghai Jiao Tong University \\
\normalsize$^3$ East China Normal University 
\quad $^4$ Peking University \\ 
\normalsize\texttt{\{shunliwang,yandanzhao,ericshding\}@tencent.com, xinyuzhourain@sjtu.edu.cn}
}

\usepackage{amssymb}
\usepackage{multirow}

\begin{document}

\maketitle

\vspace{-20pt}
\begin{figure*}[h]
    \centering
    \includegraphics[width=1.0\linewidth]{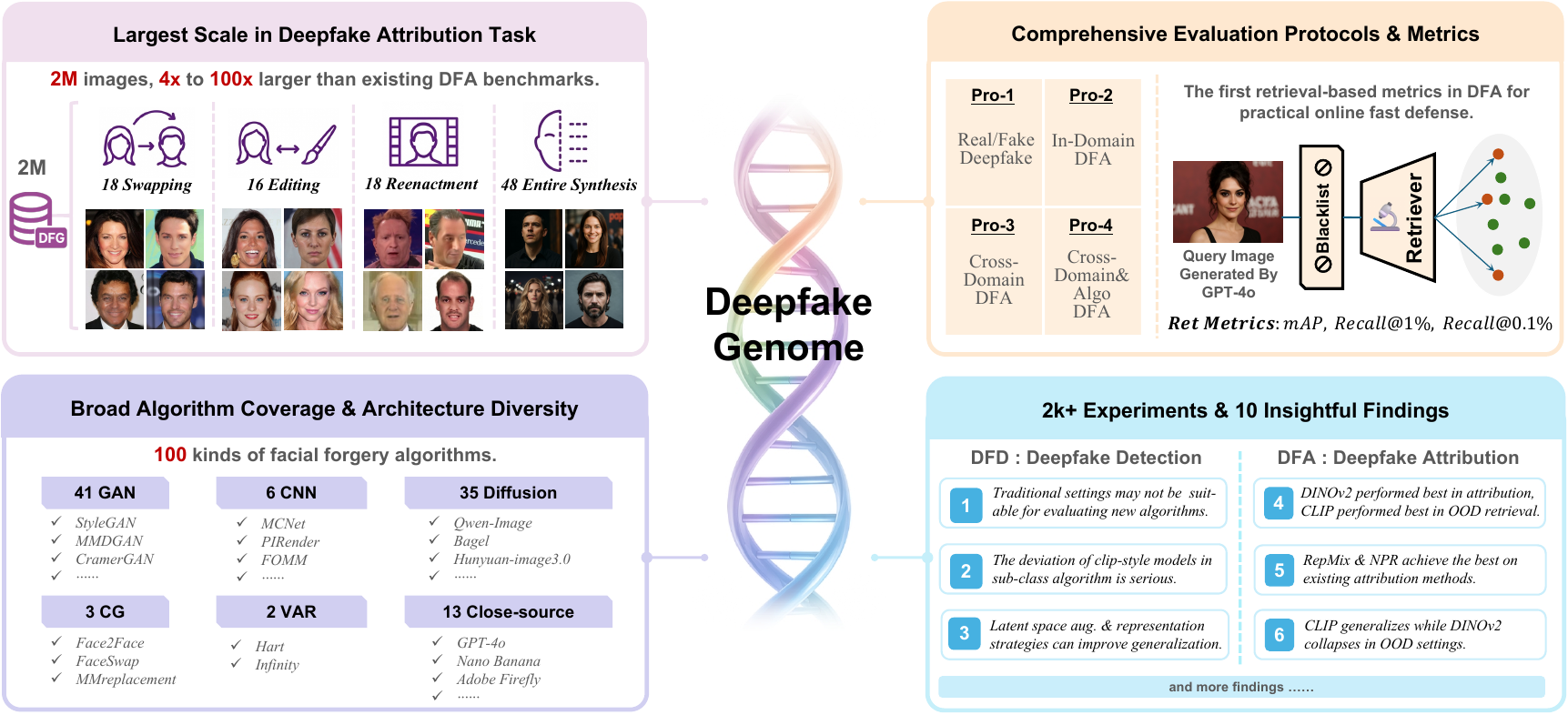}
    \caption{Overview of the proposed DeepfakeGenome benchmark. The DFG contains \textbf{100} distinct facial forgery algorithms and \textbf{2M} images. We design \textbf{4} practical evaluation protocols in Deepfake detection and attribution tasks. Over \textbf{2,000} experiments are conducted, yielding \textbf{10} insightful findings.}
    \label{fig:placeholder}
\end{figure*}

\begin{abstract}
In recent years, AIGC technologies are capable of generating hyper-realistic forged facial images, posing severe threats to facial security safety.
To tackle these challenges, early Deepfake research primarily centered on real-fake classification. Recently, growing efforts have been devoted to the Deepfake Attribution (DFA) of generated content.
Nevertheless, existing research on Deepfake detection and attribution faces saturated performance in binary classification, limited diversity in datasets and algorithms, and imperfect evaluation protocols, which severely impede practical application.
To address these limitations, we propose a comprehensive deepfake detection and attribution benchmark named \textbf{DeepfakeGenome} (DFG).
It contains \textbf{100} facial forgery algorithms and \textbf{2M} images in total, achieving \textbf{4×} to \textbf{100×} larger than prior DFA benchmarks.
We further design \textbf{4} protocols for practical evaluation, including a novel retrieval-based attribution paradigm. 
Unlike previous open-set evaluation metrics, the proposed retrieval metrics are more aligned with the real-world active defense situation of blacklist registration mechanisms.
Based on these elaborate designs, we investigate the performance ceiling of deepfake attribution task.
Over \textbf{2k} experimental evaluations are conducted, and \textbf{10} insightful findings are derived.
We believe this work can provide new insights into the DFA research field.
The Code and Benchmark of DFG are avaliable on \url{https://huggingface.co/spaces/DeepfakeGenome/DFG\_Page}.

\end{abstract}

\section{Introduction}
With the rapid advancement of deep learning, AIGC techniques are capable of generating high-fidelity content within an extremely short period \cite{Qwen-Image, nanobanana}.
These generative technologies have demonstrated remarkable value in artistic and media creation. Nevertheless, they also pose severe risks to facial security and identity authentication community.
Traditional facial Deepfake detection research \cite{Effort, DF40} focused on identifying forged contents including face swapping, facial reenactment, and facial manipulation.
With the rapid development of forgery techniques, researchers have moved beyond the binary detection task, and further focus on tracing the source of forged data to enable targeted defense.
To address this challenge, Deepfake Attribution (DFA) techniques have emerged. 
Numerous prior studies \cite{DNA-Det, universal, lorax, occ, npr, hifi, repmix, attribution} have dedicated efforts to exploring model-level source attribution for forgery images.
By investigating existing research on Deepfake detection and DFA tasks, we identify three key challenges prevailing in this field.
\begin{itemize}
    \item \textbf{Performance saturation of the binary classification on DFD benchmarks}. 
    Current ACC and AUC metrics in Deepfake detection tasks have approached saturation, indicating that existing evaluation results are likely unobjective and incomplete.
    \item \textbf{Low diversity and limited scale of DFA benchmark}. 
    Current DFA benchmarks typically include fewer than 20 algorithms, resulting in simplified, toy-level models. Consequently, there remains a substantial gap between existing methods and practical deployment.
    \item \textbf{Limitations of DFA assessment metrics}. Currently, DFA tasks usually adopt reject class classification to implement Out-of-Distribution (OOD) detection. 
    However, in the real-world attack and defense, the update iteration of forgery algorithm is extremely fast. 
    Traditional DFA metrics cannot meet this evaluation requirement.
\end{itemize}

An intuitive solution is to construct a more comprehensive and large-scale benchmark.
We conduct thorough integration and deduplication on existing Deepfake and DFA datasets \cite{FF++, DF40, Forgerynet, DFFD, DNA-Det, OSMA, wild, DiFF, Diffusionface}, and incorporate a series of recently released open-source and closed-source image forgery models \cite{Qwen-Image, BAGEL, HunyuanImage-2.1, HunyuanImage-3.0, Infinity, Hart, nanobanana, gpt-4o}. 
Ultimately, we construct a benchmark covering \textbf{100} facial forgery algorithms. 
As illustrated in Tab.~\ref{tab:compare} and Fig.~\ref{fig:compare}, the proposed DFG benchmark comprehensively outperforms previous DFA datasets in terms of both data scale and algorithm diversity.
In terms of forgery categories, DFG covers four mainstream manipulation types: \textit{Face Swapping}, \textit{Face Reenactment}, \textit{Face Editing}, and \textit{Entire Face Synthesis}.
In terms of algorithm architectures, DFG covers \textit{GAN}, \textit{Diffusion}, \textit{CNN}, \textit{CG}, and \textit{VAR} frameworks, as well as \textit{Closed-source} models.
The total number of images reaches \textbf{2M}.
As shown in Tab.~\ref{tab:compare}, the scale of the DFG dataset is \textbf{4}\textbf{$\times$} to \textbf{100}\textbf{$\times$} larger than that of existing DFA datasets.
Sufficient data scale provides a more reliable benchmark for the exploration of DFA algorithms.

Considering the practical deployment of DFA models in real-world scenarios, we design \textbf{4} \textbf{Protocols} in DFG benchmark to comprehensively assess model performance across diverse tasks.
It covers real-fake classification evaluation, deepfake attribution assessment, and retrieval-based attribution assessment under different dataset partitioning strategies.
Notably, we firstly adopt retrieval-based metrics in clustering tasks \cite{clusteringmAP} for DFA performance assessment.
This paradigm is well-suited for real-world deployment. When confronted with emerging OOD attacks, efficient defense and interception can be realized by incorporating a small number of fake samples into the flexible blacklist.

Based on the sufficient data scale and effective evaluation protocols, we conduct extensive experiments and summarize \textbf{10} highly insightful findings.
After expanding the data volume and algorithm richness, we were surprised to find that some previous consensus in Deepfake detection and DFA research was broken. 
For example, the SOTA myth of fine-tuning CLIP-ViTs \cite{CLIP} does not continue, and the improvement brought by contrastive learning is less significant than expected.
We thoroughly investigate the above phenomena via qualitative and quantitative evaluations with detailed analysis.

\section{Related Work}
\textbf{Deepfake Image Detection}.
Traditional Deepfake detection studies \cite{dct, li2021frequency, SBI, zhao2021learning, zhu2021face, khalid2020oc, larue2023seeable} primarily focus on improving the generalization performance of models against unseen forgery attacks. 
Deepfake methods can be broadly categorized into two paradigms: \textit{Learning Fake Artifacts} and \textit{Learning Real Distributions}.
The former paradigm includes frequency-based methods \cite{dct, li2021frequency}, blending-based methods \cite{SBI, zhao2021learning}, and reconstruction-based methods \cite{zhu2021face}. 
These methods attempt to model and learn inherent forgery patterns to acquire cross-attack generalization capability. 
In contrast, the latter paradigm focuses on learning from real samples, which reformulates Deepfake detection as an anomaly detection task \cite{khalid2020oc, larue2023seeable}.
However, with the continuous scaling of model sizes, existing benchmarks including FF++ \cite{FF++}, CDF-v2 \cite{cdfv2}, and DFDC \cite{DFDC} have gradually reached performance saturation. 
Therefore, the deepfake task needs more comprehensive evaluation benchmarks to achieve objective and holistic assessments of detection models.

\textbf{Synthetic Image Detection}.
Unlike Deepfake detection \cite{npr, DDA, UniFD, Effort} that only focuses on facial manipulation scenarios, general AIGC detection extends the research scope of synthetic image detection to more diverse and generalized visual contents.
These works can be divided into two categories based on whether adopting pre-trained foundational models \cite{CLIP, dinov2, dinov3}.
Some traditional methods \cite{npr, DDA} identify synthetic images by modeling forgery artifacts.
For instance, NPR \cite{npr} achieves superior generalization performance by modeling artifacts introduced by up-sampling operations.
DDA \cite{DDA} leverages VAE \cite{VAE} reconstruction to achieve strict semantic alignment between real and fake samples, thereby eliminating inherent biases during the learning process.
In comparison, other methods \cite{UniFD, Effort} exploit the pre-trained knowledge in CLIP-ViT \cite{CLIP} and perform diverse fine-tuning strategies to detect fake images.

\textbf{Deepfake Attribution}.
Different from naive real/fake classification, DeepFake Attribution (DFA) task identifies the specific generation algorithm of fake images, providing more comprehensive analytical cues for forgery defense strategies.
Early studies \cite{attributing, marra2019dogans, repmix, xu2025detecting} primarily focus on exploring inherent fingerprint characteristics of GAN models.
Recent research \cite{OWDFA++, POSE, li2021frequency} has further extended deepfake attribution to open-world scenarios. 
Sun \textit{et al.} \cite{OWDFA++} introduced OWDFA benchmark, which aims to perform attribution on semi-supervised setting.
Yang \textit{et al.} \cite{POSE} propose POSE framework to attribute images to known models and identify those from unknown ones.
Li \textit{et al.} \cite{li2021frequency} integrated high-pass filters and CLIP-ViT \cite{CLIP} to capture fine-grained forgery features to perform framework attribution based on reference sets.
However, these methods are trained and evaluated on limited data scale and few algorithms. There remains a substantial gap between these models and practical applications.


\section{DeepfakeGenome Benchmark}
\label{headings}

\textbf{Brief introduction of DFG Benchmark}. We systematically integrate and deduplicate existing deepfake and DFA datasets (FF++ \cite{FF++}, DF40 \cite{DF40}, ForgeryNet \cite{Forgerynet}, DFFD \cite{DFFD}, DNA-Det \cite{DNA-Det}, OSMA \cite{OSMA}, Wild-20 \cite{wild}, DiFF \cite{DiFF}, DiffusionFace \cite{Diffusionface}), and sorted out 92 deepfake algorithms.
After that, we synthesized 8 kinds of recently accessible Text-to-Image algorithms: Qwen-Image \cite{Qwen-Image}, BAGEL \cite{BAGEL}, HunyuanImage-2.1/3.0 \cite{HunyuanImage-2.1, HunyuanImage-3.0}, Infinity \cite{Infinity}, Hart \cite{Hart}, Nanao Banana \cite{nanobanana} and GPT-4o \cite{gpt-4o}. As shown in \textbf{Tab.~\ref{tab:DFG-100-algos}}, the DeepfakeGenome benchmark includes \textbf{100} kinds of face forgery algorithms.
The total number of frames of the DFG has reached \textbf{2M}.
Tab.~\ref{tab:compare} and Fig.~\ref{fig:compare} compare the DFG benchmark with other DFA benchmarks. It surpasses the existing DFA datasets in the diversity of algorithms, the volume of frames and the novelty of forgery algorithms.

\begin{figure}[h]
    \centering
    \includegraphics[width=0.95\linewidth]{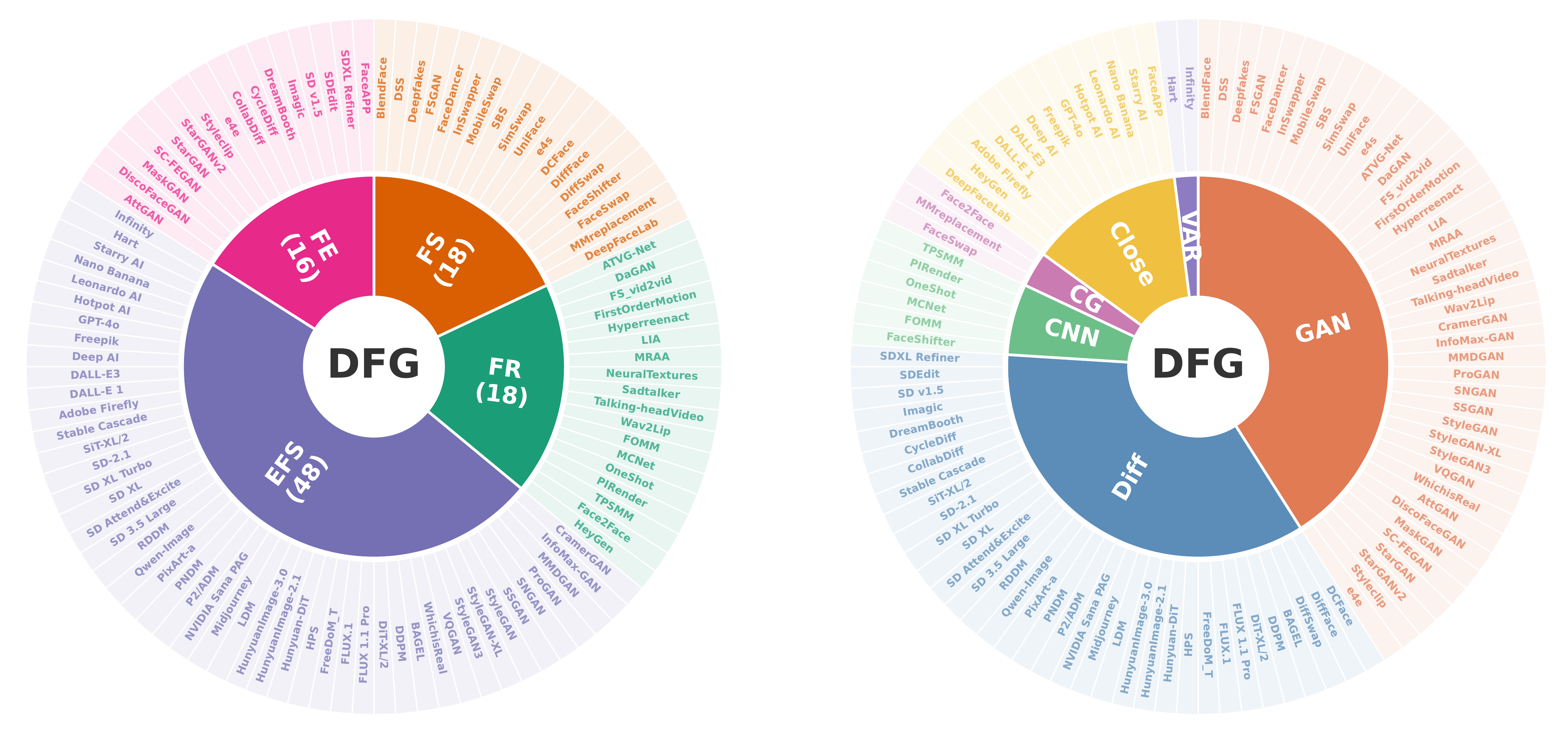}
    \caption{Classify 100 deepfake algorithms in terms of forgery types and algorithm architectures.}
    \label{fig:cake}
\end{figure}

Fig.~\ref{fig:cake} demonstrates the distribution of 100 algorithms on two taxonomy.
In terms of the forgery types, DFG contains 4 forgery types: \textbf{18} \textit{Face Swapping} (FS), \textbf{18} \textit{Face  Reenactment} (FR), \textbf{48} \textit{Entire Face Synthesis} (EFS), and \textbf{16} \textit{Face Editing} (FE) algorithms.
In algorithm architectures, DFG contains \textbf{6} classic \textit{Convolutional Neural Network} (CNN), \textbf{3} \textit{Computer Graphics} (CG), \textbf{41} \textit{Generative Adversarial Network} (GAN), \textbf{35} \textit{Diffusion}, \textbf{13} \textit{Close-source}, and \textbf{2} \textit{Visual Autoregressive} (VAR) algorithms.

\begin{table}[htbp]
    \centering
    \begin{minipage}[c]{0.49\textwidth}
        \centering
        \caption{A brief summary of the existing DFA datasets and the DeepfakeGenome dataset.}
        \label{tab:compare}         
        \scriptsize
        \setlength{\tabcolsep}{2pt}
        \begin{tabular}{l c l c r}
            \toprule
            \textbf{DFA Dataset} & \textbf{Venue} & \textbf{Data Domain} & \textbf{\#Alg.} & \textbf{\#Fms} \\
            \midrule
            GAN Finger. \cite{attributing}    & ICCV 2019  & Nature\&Face & 4  & 110k \\
            DFA \cite{attribution}            & ArXiv 2021 & Face         & 3  & 18k \\
            OWGAN \cite{towards}              & ICCV 2021  & Nature\&Face & 11 & -- \\
            Attribution88 \cite{repmix}       & ECCV 2022  & Face         & 7  & 105k \\
            DNA-Det \cite{deepfake}           & AAAI 2022  & Nature\&Face & 4  & 509k \\
            OSMA \cite{progressive}           & CVPR 2023  & Nature\&Face & 17 & 281k \\
            OWDFA \cite{contrastive}          & ICCV 2023  & Face         & 20 & 59k \\
            DEFL \cite{handcrafted}           & MM 2024    & Nature\&Face & 12 & 26k \\
            OWDFA++ \cite{rethinking}         & IJCV 2024  & Face         & 25 & 338k \\
            OSS-DFA \cite{data}               & ArXiv 2025 & Face         & 20 & -- \\
            WILD-20 \cite{wild}               & arXiv 2025 & Face         & 20 & 20k \\
            BMRL \cite{pvlm}                  & ArXiv 2026 & Face         & 28 & 135k \\
            \midrule
            DFG (Ours)                        & 2026       & Face         & 100& 2M \\
            \bottomrule
        \end{tabular}
    \end{minipage}
    \hfill
    \begin{minipage}[c]{0.49\textwidth}
        \centering
        \includegraphics[width=\linewidth]{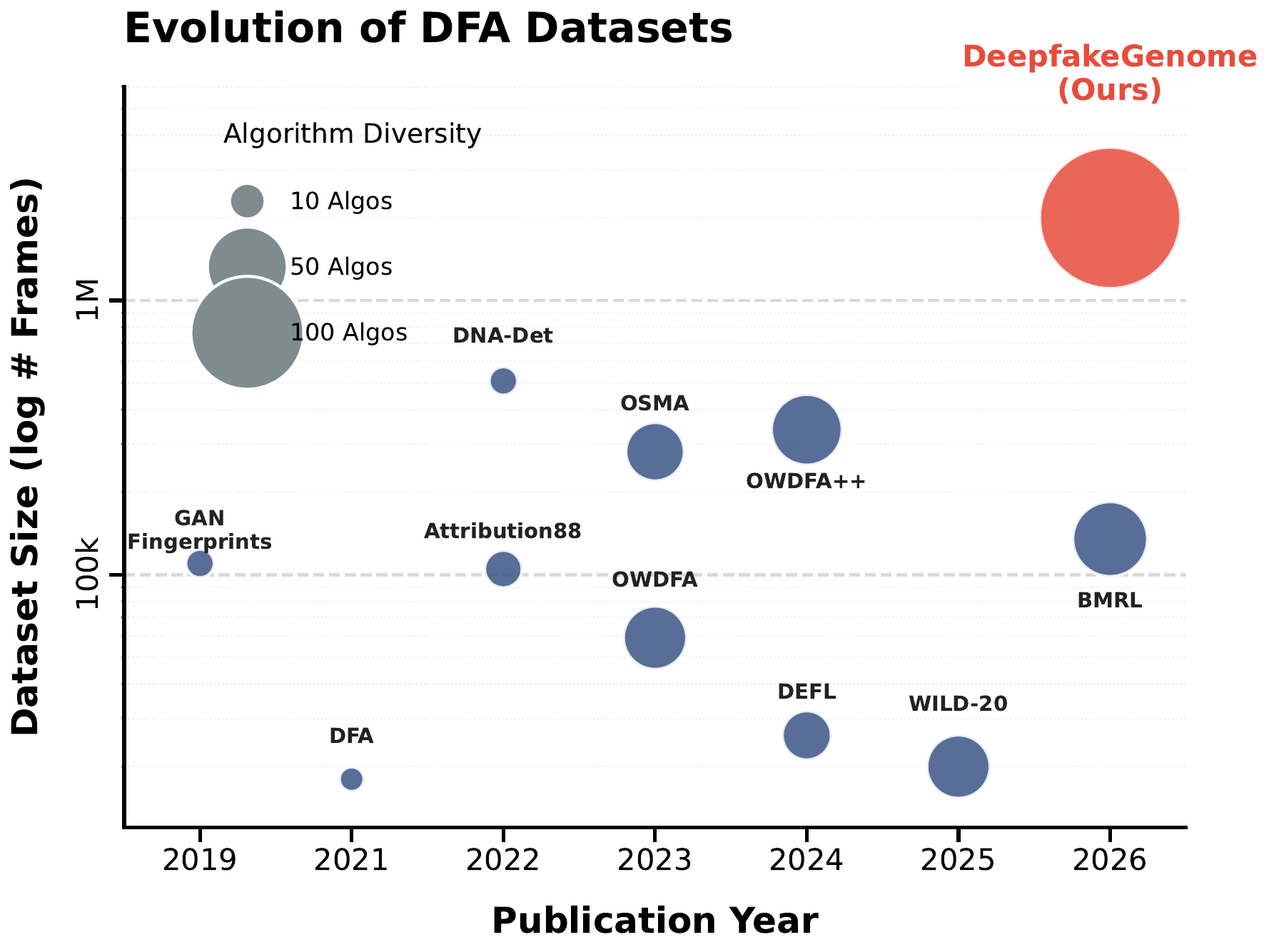}
        \captionof{figure}{Volume comparison between the DeepfakeGenome and previous DFA datasets.}
        \label{fig:compare}         
    \end{minipage}
    \vspace{-1em}
\end{table}

\begin{figure}[h]
    \centering
    \includegraphics[width=1.0\linewidth]{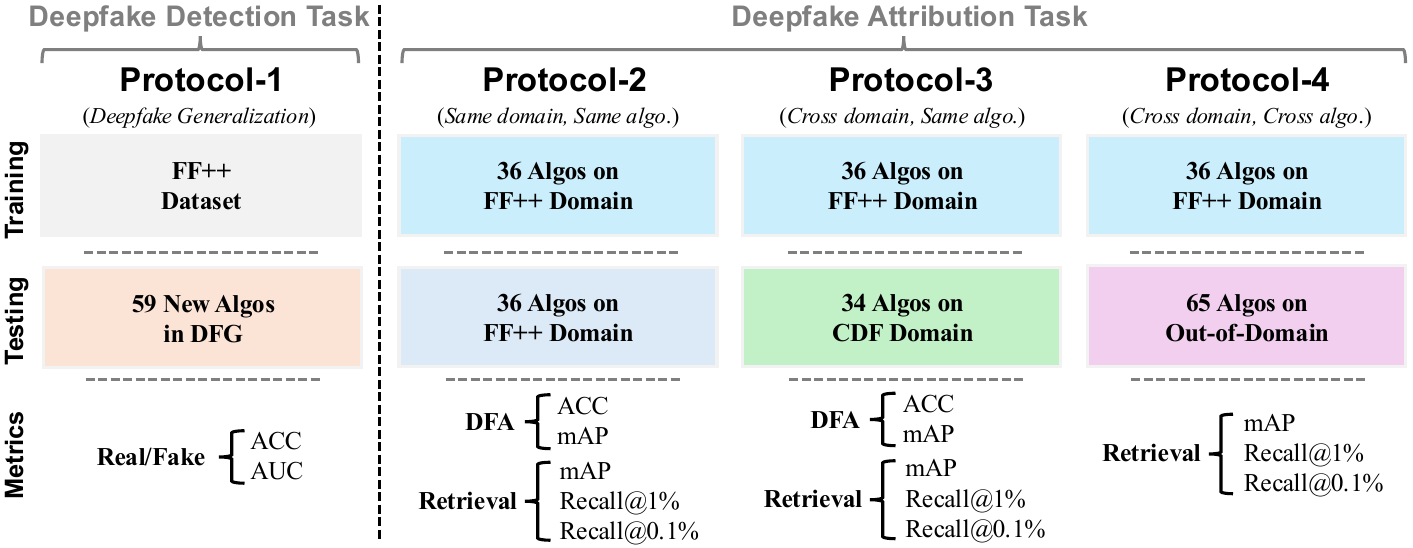}
    \caption{Four data partitioning protocols of the DFG Benchmark.}
    \label{fig:4-protocol}
\end{figure}

\textbf{Data Partitioning Protocols}. 
To fully exploit the abundant data resources contained in DFG, we designed 4 data partitioning protocols to conduct experiments.
Fig.~\ref{fig:4-protocol} distinguishes 4 protocols with different colors.
Tab.~\ref{tab:pro-1-traintest}-\ref{tab:pro-4-traintest} in the supplementary material lists the specific algorithms and data domains in each partition in detail.
The overall design criteria are: data in FF++ \cite{FF++} domain is for training, and data in other domains is for performance evaluation.
In \textbf{Protocol-1}, we follow the traditional deepfake settings \cite{Effort}: training on vanilla FF++ \cite{FF++} dataset (DeepFakes \cite{Deepfakes}, Face2Face \cite{Face2face}, FaceSwap \cite{FaceSwap} and NeuralTextures \cite{NeuralTextures}), evaluating ACC and AUC on 59 new algorithms of DFG (note that all algorithms in DF40 \cite{DF40} are removed because the performance has been explored in DF40 \cite{DF40} paper).
In \textbf{Protocol-2} \& \textbf{Protocol-3}, 36 algorithms generated in FF++ domain are adopted as the training set, 36 and 34 algorithms in FF++ and CDF domain are adopted as the testing set, respectively.
In this way, \textit{Same Domain, Same Algorithm} and \textit{Cross Domain, Same Algorithm} settings can be constructed.
In \textbf{Protocol-4}, we adopt 36 algorithms in FF++ domain as training set, and conduct retrieval test for 65 algorithms out of FF++ domain and 36 training algorithms. 
The \textit{Cross Domain, Cross Algorithm} setting is most suitable for real-world application scenarios.

\textbf{Performance Metrics}. 
In \textbf{Protocol-1}, we adopt ACC and AUC in the classic deepfake task \cite{DF40, Effort} as performance metrics;
In \textbf{Protocol-2} and \textbf{Protocol-3}, since the forgery algorithms in the training and testing sets are identical, the variation lies solely in the data domain. 
First, we adopt \textit{Accuracy} (\textit{ACC}) and \textit{mean Average Precision} (\textit{mAP}) from multi-class classification tasks as our primary attribution performance metrics. 
Furthermore, to rigorously assess the quality of the feature representations extracted by the model, we introduce \textit{Ret-mAP} and \textit{Ret-Recall} from information retrieval tasks as additional evaluation metrics. 
Specifically, suppose there are $M$ forgery algorithms in total, and $K_j$ samples belong to the $j$-th class. In the retrieval setting, \textit{Ret-mAP} measures the model's capability to rank relevant samples highly in the retrieved list:
\begin{equation}
    AP_j = \frac{1}{K_j} \sum_{i=1}^{K_j} \frac{i}{Rank_j(i)}, \quad mAP = \frac{1}{M} \sum_{j=1}^{M} AP_{j},
\end{equation}
where $Rank_j(i)$ denotes the rank position of the $i$-th relevant sample of the $j$-th class after retrieval and sorting.
To calculate the \textit{Ret-Recall}, we first randomly sample 10 query instances from a specific class. We then extract their features and compute the average to obtain a representative query feature vector, denoted as $f_{query}$. 
Subsequently, we establish a strict threshold $\tau_{\alpha}$ based on \textit{False Rejection Rate} (\textit{FRR}), denoted as $\alpha$ (\textit{e.g.}, $\alpha = 1\%$ or $0.1\%$). 
Under this threshold, we measure the proportion of relevant negative samples (DeepFakes) that are successfully recalled. The formula for $Recall@\alpha$ is defined as:
\begin{equation}
    Recall@\alpha = \frac{K_{recall}}{K_j}, \quad K_{recall} = \sum_{x \in \mathcal{S}_j} \mathbb{I} \big( sim(f_{query}, f_x) \ge \tau_{\alpha} \big),
\end{equation}
where $\mathcal{S}_j$ represents the set of samples in the $j$-th class (with $|\mathcal{S}_j| = K_j$), $f_x$ is the feature of sample $x$, $sim(\cdot, \cdot)$ denotes the similarity metric (\textit{e.g.}, cosine similarity), and $\mathbb{I}(\cdot)$ is the indicator function that equals 1 if the condition is met and 0 otherwise.
In \textbf{Protocol-4}, we adopt \textit{Ret-mAP}, \textit{Recall@1\%} and \textit{Recall@0.1\%} to measure the dynamic interception performance against unseen forgery algorithms.
Note that these metrics are particularly well-suited for \textbf{Blacklist Interception} scenarios in real-world applications, enabling a rapid and robust response to novel attacks.

\section{Experiments \& Findings}
\label{others}

\subsection{Experiment Details}
During training, we use the Adam optimizer. The learning rate is set to
$2 \times 10^{-5}$ for fine-tuning CLIP-ViT \cite{CLIP} models and to $2 \times 10^{-4}$
for all other models. The batch size is 32 for both training and testing,
and all models are trained for 20 epochs. All CLIP-ViT \cite{CLIP} models are initialized
with the default CLIP pre-trained weights released by OpenAI. 
For the ViT \cite{ViT}, CLIP \cite{CLIP}, and DINO \cite{dinov2} model families, input images are resized to $224 \times 224$ pixels. For CNN-based models, images are resized to $256 \times 256$ pixels. When data augmentation is enabled, the pipeline includes
\texttt{HorizontalFlipping} with probability $0.5$;
\texttt{RandomBrightnessContrast} with limits of $0.2$ and probability
$0.5$; \texttt{HueSaturationValue} with probability $0.3$; and
\texttt{ImageCompression} with quality sampled from $[40, 100]$ and
probability $0.1$. \texttt{GaussianNoise}, \texttt{MotionBlur},
\texttt{CLAHE}, \texttt{ChannelShuffle}, and \texttt{Cutout} are each
applied with probability $0.1$, while \texttt{RandomGamma} and
\texttt{GlassBlur} are each applied with probability $0.3$. All experiments are conducted using an Intel Xeon Platinum 8255C CPU @ 2.50\,GHz and four
NVIDIA V100 GPUs with 32GB.

\subsection{Deepfake Detection Results (Pro-1)}
\begin{table*}[htbp]
    \centering
    \caption{ACC \& AUC on 59 algorithms in terms of algorithm types. The \textbf{highest} and \underline{second} performance of each row are highlighted.}
    \label{tab:pro1-types}
    \footnotesize  
    \renewcommand{\arraystretch}{1.0}  
        \resizebox{1.0\textwidth}{!}{  
        \begin{tabular}{l|cc|cc|cc|ccc|cc|ccc|c} 
        \toprule
        \multirow{4}{*}{\textbf{Types}} & \multicolumn{2}{c|}{\textbf{Basic Backbone}} & \multicolumn{2}{c|}{\textbf{Finetuned CLIP}} & \multicolumn{2}{c|}{\textbf{Reconstruction-based}} & \multicolumn{3}{c|}{\textbf{Frequency-domain}} & \multicolumn{2}{c|}{\textbf{Latent Space}} & \multicolumn{3}{c|}{\textbf{Representation}} & \multicolumn{1}{c}{\textbf{Aug Adv.}} \\ 
        \cmidrule{2-3} \cmidrule{4-16}  
         & \textbf{Xception} & \textbf{Eff-B4} & \textbf{CLIP-ViT-FFT} & \textbf{Effort} & \textbf{RECCE} & \textbf{SBI} & \textbf{F3Net} & \textbf{SRM} & \textbf{SPSL} & \textbf{ProDet} & \textbf{LSDA} & \textbf{UCF} & \textbf{IID} & \textbf{CORE} & \textbf{sladd} \\
        & \cite{xception} & \cite{EfficientNet} & \cite{CLIP} & \cite{Effort} & \cite{recce} & \cite{SBI} & \cite{F3Net} & \cite{SRM} & \cite{SPSL} & \cite{ProDet} & \cite{LSDA} & \cite{UCF} & \cite{iid} & \cite{CORE} & \cite{sladd} \\ 
        \midrule
        FR (3 Algos)     & 0.6854 & 0.7144 & \underline{0.7323} & \textbf{0.8252} & 0.6976 & 0.6937 & 0.7073 & 0.6644 & 0.6157 & 0.6801 & 0.7189 & 0.7157 & 0.6767 & 0.7101 & 0.7033 \\
        FS (6 Algos)     & 0.6416 & 0.6637 & \underline{0.7752} & \textbf{0.8184} & 0.6194 & 0.6098 & 0.6088 & 0.6121 & 0.6465 & 0.6167 & 0.6128 & 0.6488 & 0.6087 & 0.6393 & 0.6057 \\
        FE (11 Algos)    & 0.5613 & 0.5939 & \underline{0.6315} & \textbf{0.7586} & 0.5762 & 0.5897 & 0.5908 & 0.5846 & 0.5643 & 0.5603 & 0.5908 & 0.5908 & 0.6080 & 0.6032 & 0.6051 \\
        EFS (39 Algos)   & 0.5923 & 0.5486 & 0.5664 & 0.5895 & 0.6052 & 0.6123 & 0.5682 & 0.5821 & 0.5445 & 0.5021 & \textbf{0.6841} & \underline{0.6841} & 0.6487 & 0.6398 & 0.5962 \\
        Avg.            & 0.5962 & 0.5772 & 0.6082 & \underline{0.6563} & 0.6060 & 0.4709 & 0.6127 & 0.5898 & 0.5621 & 0.5337 & \textbf{0.6612} & 0.5881 & 0.6385 & 0.6365 & 0.6043 \\
        \midrule
        FR (3 Algos)     & 0.7680 & 0.8475 & \underline{0.8622} & \textbf{0.9343} & 0.7722 & 0.7813 & 0.7959 & 0.7565 & 0.7666 & 0.8306 & 0.8027 & 0.8027 & 0.7511 & 0.7930 & 0.8015 \\
        FS (6 Algos)     & 0.7056 & 0.7249 & \underline{0.8610} & \textbf{0.9017} & 0.6958 & 0.6997 & 0.7403 & 0.6728 & 0.7175 & 0.6614 & 0.6856 & 0.6856 & 0.6554 & 0.7155 & 0.6533 \\
        FE (11 Algos)    & 0.6493 & 0.7026 & \underline{0.7370} & \textbf{0.8420} & 0.6647 & 0.6988 & 0.7037 & 0.6434 & 0.6330 & 0.5926 & 0.6545 & 0.6545 & 0.6632 & 0.7282 & 0.6436 \\
        EFS (39 Algos)   & 0.6316 & 0.7400 & 0.5983 & 0.5737 & 0.6642 & 0.6830 & 0.5978 & 0.6448 & 0.5535 & 0.5139 & \textbf{0.7511} & \underline{0.7511} & 0.7028 & 0.7081 & 0.6092 \\
        Avg.            & 0.6494 & 0.6119 & 0.6643 & 0.6754 & 0.6730 & 0.4645 & 0.6926 & 0.6531 & 0.5959 & 0.5597 & \textbf{0.7291} & 0.6421 & 0.6931 & \underline{0.7169} & 0.6299 \\
        \bottomrule
        \end{tabular}
        }%
    \end{table*}

\begin{table*}[htbp]
    \centering
    \caption{ACC \& AUC on 59 algorithms in terms of algorithm architectures.}
    \label{tab:pro1-architectures}
    \footnotesize  
    \renewcommand{\arraystretch}{1.0}  
        \resizebox{1.0\textwidth}{!}{  
        \begin{tabular}{l|cc|cc|cc|ccc|cc|ccc|c}  
        \toprule
        \multirow{4}{*}{\textbf{Architecture}} & \multicolumn{2}{c|}{\textbf{Basic Backbone}} & \multicolumn{2}{c|}{\textbf{Finetuned CLIP}} & \multicolumn{2}{c|}{\textbf{Reconstruction-based}} & \multicolumn{3}{c|}{\textbf{Frequency-domain}} & \multicolumn{2}{c|}{\textbf{Latent Space}} & \multicolumn{3}{c|}{\textbf{Representation}} & \multicolumn{1}{c}{\textbf{Aug Adv.}} \\ 
        \cmidrule{2-16} 
        & \textbf{Xception} & \textbf{Eff-B4} & \textbf{CLIP-ViT-FFT} & \textbf{Effort} & \textbf{RECCE} & \textbf{SBI} & \textbf{F3Net} & \textbf{SRM} & \textbf{SPSL} & \textbf{ProDet} & \textbf{LSDA} & \textbf{UCF} & \textbf{IID} & \textbf{CORE} & \textbf{sladd} \\
        & \cite{xception} & \cite{EfficientNet} & \cite{CLIP} & \cite{Effort} & \cite{recce} & \cite{SBI} & \cite{F3Net} & \cite{SRM} & \cite{SPSL} & \cite{ProDet} & \cite{LSDA} & \cite{UCF} & \cite{iid} & \cite{CORE} & \cite{sladd} \\ 
        \midrule
        CG (1 Algo)          & 0.5252 & 0.5430 & \textbf{0.5837} & \underline{0.5824} & 0.4954 & 0.4823 & 0.5072 & 0.5006 & 0.5359 & 0.4819 & 0.4863 & 0.5064 & 0.4751 & 0.5206 & 0.4821 \\
        GAN (16 Algos)       & 0.5975 & 0.6630 & \underline{0.7352} & \textbf{0.8477} & 0.6157 & 0.6135 & 0.6430 & 0.5806 & 0.5894 & 0.5608 & 0.6395 & 0.6607 & 0.6122 & 0.6366 & 0.5923 \\
        Diffusion (29 Algos) & 0.5874 & 0.5396 & 0.5710 & 0.6141 & 0.5926 & 0.4309 & 0.5865 & 0.5932 & 0.5665 & 0.5556 & \underline{0.6440} & 0.5568 & \textbf{0.6479} & 0.6339 & 0.6114 \\
        Close (11 Algos)     & 0.6161 & 0.5508 & 0.5393 & 0.5153 & 0.6277 & 0.3876 & 0.6433 & 0.5983 & 0.5169 & 0.4464 & \textbf{0.7383} & 0.5757 & \underline{0.6573} & 0.6509 & 0.6146 \\
        VAR (2 Algos)        & 0.6413 & 0.5981 & 0.5216 & 0.5491 & 0.6579 & 0.6336 & 0.6336 & 0.6109 & 0.5436 & 0.5050 & \textbf{0.7475} & 0.5704 & \underline{0.6900} & 0.6526 & 0.6002 \\
        Avg.            & 0.5962 & 0.5772 & 0.6082 & \underline{0.6563} & 0.6060 & 0.4709 & 0.6127 & 0.5898 & 0.5621 & 0.5337 & \textbf{0.6612} & 0.5881 & 0.6385 & 0.6365 & 0.6043 \\
        \midrule
        CG (1 Algo)          & 0.5622 & 0.5732 & \underline{0.6284} & \textbf{0.6336} & 0.5038 & 0.4873 & 0.5467 & 0.5026 & 0.5384 & 0.4746 & 0.4908 & 0.5493 & 0.4624 & 0.5562 & 0.4672 \\
        GAN (16 Algos)       & 0.6571 & 0.7299 & \underline{0.8437} & \textbf{0.9574} & 0.6995 & 0.6559 & 0.7477 & 0.6453 & 0.6960 & 0.6004 & 0.6956 & 0.7556 & 0.6567 & 0.7473 & 0.6045 \\
        Diffusion (29 Algos) & 0.6404 & 0.5708 & 0.6229 & 0.6194 & 0.6638 & 0.4231 & 0.6658 & 0.6501 & 0.5715 & 0.5767 & \textbf{0.7168} & 0.6196 & 0.7038 & \underline{0.7098} & 0.6404 \\
        Close (11 Algos)     & 0.6666 & 0.5508 & 0.5418 & 0.4410 & 0.6676 & 0.3224 & 0.7016 & 0.6798 & 0.5281 & 0.4692 & \textbf{0.8147} & 0.5521 & \underline{0.7249} & 0.7116 & 0.6518 \\
        VAR (2 Algos)        & 0.6661 & 0.6168 & 0.5199 & 0.5418 & 0.7096 & 0.3033 & 0.6646 & 0.6871 & 0.5486 & 0.5276 & \textbf{0.8221} & 0.6018 & \underline{0.7684} & 0.6864 & 0.6411 \\
        Avg.            & 0.6494 & 0.6119 & 0.6643 & 0.6754 & 0.6730 & 0.4645 & 0.6926 & 0.6531 & 0.5959 & 0.5597 & \textbf{0.7291} & 0.6421 & 0.6931 & \underline{0.7169} & 0.6299 \\
        \bottomrule
        \end{tabular}
        }%
    \end{table*}
    
\begin{figure*}[htbp]
    \centering
    \includegraphics[width=1.0\linewidth]{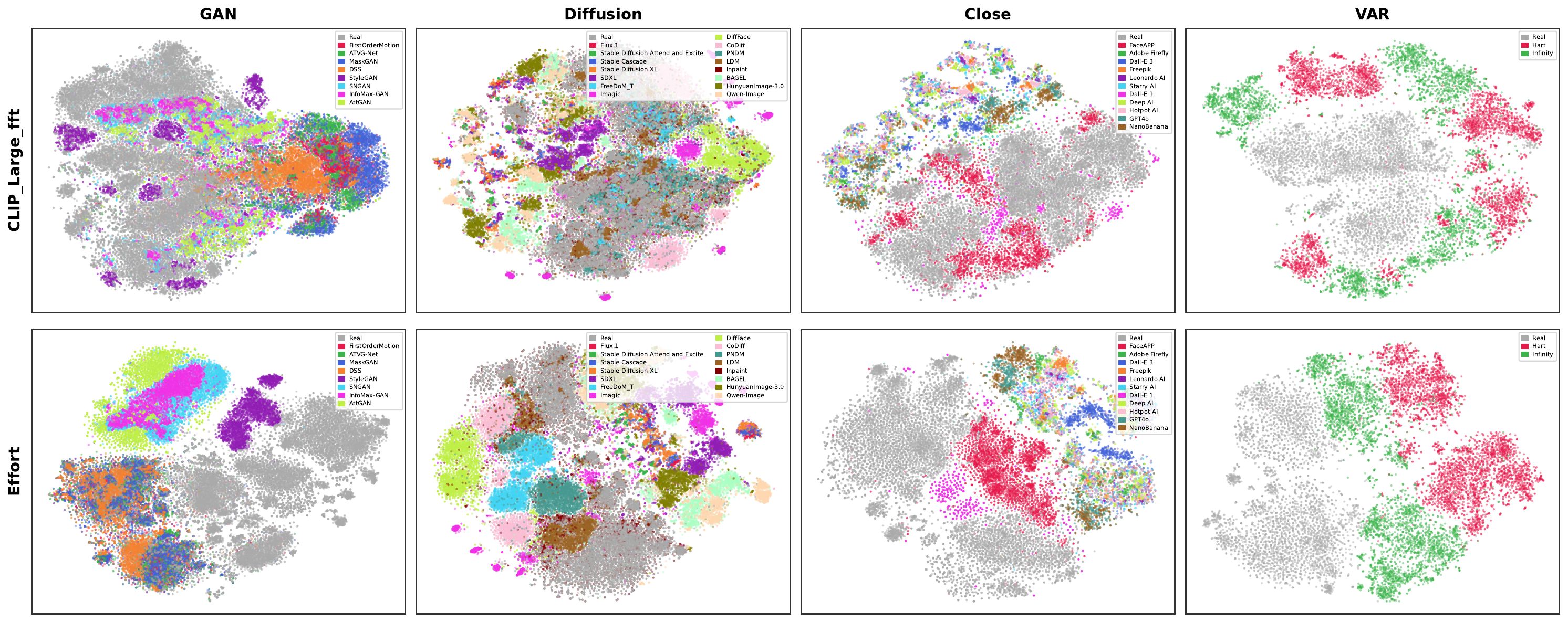}
    \caption{t-SNE visualization of CLIP-ViT-FFT \cite{CLIP} and Effort \cite{Effort} features on GAN, Diffusion, Close-source and VAR architectures.}
    \label{fig:pro-1-tSNE}
\end{figure*}

\textbf{Finding-1}. \textbf{Traditional Deepfake training settings may not be suitable for evaluating emerging algorithms}.
Currently, a general consensus \cite{Effort} in deepfake detection task is that fine-tuning the CLIP-ViT \cite{CLIP} model can achieve the best generalization performance.
However, as shown in Tab.~\ref{tab:pro1-types}\&\ref{tab:pro1-architectures}, the generalization results on 59 algorithms does not support this consensus.
Specifically, in terms of ACC, the average performance (0.6563) of Effort \cite{Effort} is indeed higher than CLIP-ViT-FFT \cite{CLIP} (0.6082), but is slightly lower than LSDA \cite{LSDA} (0.6612).
In terms of AUC, the overall performance of clip-style models \cite{CLIP, Effort} (0.6643, 0.6754) are worse than LSDA \cite{LSDA}, IID \cite{iid} and CORE \cite{CORE} (0.7291, 0.6931, 0.7169). 
We argue that the performance degradation of the clip-style model stems not from poor generalization of CLIP-ViT, but from the over-simplified distribution of the FF++ \cite{FF++} training dataset. Such a limited dataset causes overfitting when trained with the highly complex CLIP-ViT \cite{CLIP} backbone.
This also indicates that the existing deepfake evaluation paradigm may have already fallen behind novel forgery algorithms, and more benchmark explorations are very necessary.

\textbf{Finding-2}. \textbf{The deviation of clip-style models in sub-class algorithm is serious}.
Although the clip-style models \cite{CLIP, Effort} achieve SOTA AUC performance on FR, FS and FE, it can only achieve 0.5983 and 0.5737 in EFS category respectively. Far behind LSDA \cite{LSDA}, IID \cite{iid} and CORE \cite{CORE} (0.7511, 0.7028, 0.7081).
One possible explanation is \textit{Excessive Semantic Deviation}: these models are fine-tuned from a high-quality real semantic space to produce offset, thus separating the real image from the easy forgery images in FF++ Datasets.
However, when encountering more realistic EFS images generated by diffusion architectures, the biased decision preference that deviates from real semantics is very prone to errors.
This is confirmed by the performance statistics from architecture level in Tab.~\ref{tab:pro1-architectures}.
The clip-style models \cite{CLIP, Effort} outperforms all other models in GAN and CG frameworks, but lags far behind LSDA \cite{LSDA} and IID \cite{iid} models in the Diffusion, Close-source and VAR frameworks (the performance gap even reaches 0.2).
To qualitatively evaluate the performance of the model on different architectures, we visualize t-SNE of clip-style models in Fig.~\ref{fig:pro-1-tSNE}.
Note that some algorithms are closed considering the clarity.
Real samples are colored in gray dots for display.
It can be found that the two models can be clearly distinguished on the GAN architectures. However, there is confusion in the Diffusion architectures.



\textbf{Finding-3}. \textbf{Latent space augmentation and representation learning strategies can significantly improve the generalization on new attacks}.
LSDA \cite{LSDA} and IID \cite{iid} achieve the highest AUC performance in Diffusion, Close-source and VAR architectures respectively. 
And these performances are much higher than other models.
This demonstrates that the latent augmentation and representation learning on FF++ can effectively improve the generalization performance on new forgery architectures.
This finding has important guiding significance for future AIGC detection.
While the specific techniques are also very important. 
For example, ProDet \cite{ProDet} adopts a progressive feature augmentation strategy, while LSDA directly combines features across forgery types.


\subsection{In\&Out-of-Domain Attribution Results (Pro-2\&3)}

\begin{wraptable}{r}{0.5\textwidth} 
    \centering
    \scriptsize
    \setlength{\tabcolsep}{1.5pt}
    \begin{threeparttable}
    \caption{\small mAP \& Recall of backbones and attribution methods on Protocol-4.The default rank in LORA is 1.}
    \label{tab:backbone on Pro4}
    \begin{tabular}{l|ccc|c}
    \toprule
    \multirow{2}{*}{\textbf{Methods}} 
    &\multirow{2}{*}{\textbf{Ret-mAP}}&\textbf{Ret-Recall}&\textbf{Ret-Recall}&\multirow{2}{*}{\textbf{Avg}}\\
    & &\textbf{@1\%}&\textbf{@0.1\%}\\
    \midrule
    \multicolumn{5}{c}{\textbf{Visual BackBones}}\\
    \midrule
    Xception \cite{xception}&0.2493 & 0.4586 & 0.2916 &0.3331\\
    F3Net  \cite{F3Net}&0.2736 & 0.4289 & 0.2567 &0.3197\\
    Resnet-34 \cite{resnet}&0.2745 & 0.4242 & 0.276 &0.3249\\
    Effort  \cite{Effort}&\textbf{0.5753} & \textbf{0.8607} & \textbf{0.7727} &\textbf{0.7362}\\
    CLIP-ViT\_FFT \cite{CLIP}&\underline{0.5507} & \underline{0.8355} & \underline{0.7358} &\underline{0.7073}\\
    CLIP-ViT\_LORA \cite{lora}&0.4687 & 0.7819 & 0.6805 &0.6437\\
    DINOv2-ViT\_FFT \cite{dinov2}&0.3618 & 0.6459 & 0.4542& 0.4873\\
    DINOv3-ViT\_FFT \cite{dinov3}&0.3788 & 0.5703 & 0.4054 &0.4515\\
    \midrule
    \multicolumn{5}{c}{\textbf{Attribution Methods}}\\
    \midrule
    DCT \cite{dct}&0.2742 & 0.4997 & 0.3948 &0.3895\\
    DNA-DET \cite{deepfake}&0.1535 & 0.2151 & 0.1257 &0.1647\\
    Universal \cite{universal} &0.1974 & 0.3343 & 0.1592 &0.2303\\
    LoRAX \cite{lorax}&0.3069 & 0.5365 & 0.3635 &0.4023\\
    OCC-CLIP \cite{occ}&\underline{0.3131} & \underline{0.5482} & 0.3463 &0.4025\\
    NPR \cite{npr}&0.\textbf{3312} & 0.\textbf{6345} & 0.\textbf{4707} &\textbf{0.4788}\\
    POSE \cite{progressive}&0.2476 & 0.3522 & 0.2284 &0.2760\\
    HIFI-IFD \cite{hifi}&0.2819 & 0.5397 & \underline{0.3992} &\underline{0.4069}\\
    Repmix \cite{repmix}&0.2950 & 0.4588 & 0.2788 &0.3442\\
    GAN Attribution \cite{attributing}&0.2061 & 0.2756 & 0.1470 &0.2095\\
    \bottomrule
    \end{tabular}
    \end{threeparttable}
\end{wraptable}

\textbf{Finding-4}. \textbf{DINOv2 performed best in attribution, while CLIP performed best in OOD retrieval}.
As shown in Tab.~\ref{tab:backbone on Pro2&3}, fully fine-tuned DINOv2 \cite{dinov2} achieves the best attribution performance on Pro-2\&3 (\textit{e.g.}, \textit{ACC} and \textit{mAP}). 
This indicates that the low-level multi-crop pre-training are more effective for capturing subtle deepfake fingerprints generated by different manipulation algorithms. This phenomenon has also been reported in \cite{zhang2026fine}.
In the retrieval task, the fine-tuning CLIP-ViT \cite{CLIP} demonstrates superior performance in retrieval metrics (\textit{e.g.}, \textit{Ret-mAP} and \textit{Ret-Recall}).
This advantage stems from the strong global semantic representation ability of CLIP model, which provides a good start for the learning of model fingerprints.
One obvious phenomenon is that the Effort \cite{Effort} lags behind the fine-tuned CLIP and DINO models in terms of attribution performance.
This indicates that the conservative SVD fine-tuning strategy retains most of the semantic features in CLIP, but it does not provide much help in attribution learning.

\textbf{Finding-5}. \textbf{RepMix and NPR achieve the best performance among the existing attribution methods}.
Tab.~\ref{tab:backbone on Pro2&3} shows that RepMix \cite{repmix} and NPR \cite{npr} have achieved the best attribution performance in Protocol-2\&3 respectively. 
The RepMix \cite{repmix} adopted a latent feature mixing strategy to synthesize data from interpolation between existing samples.
Unlike the LSDA \cite{LSDA}, the mixing operation occurs between real and fake samples, not between fake and fake.
The latent space feature augmentation can significantly improve the attribution ability of the model in the same domain, but its performance is very poor in the OOD setting, as shown in Tab.~\ref{tab:backbone on Pro4}.
The core idea of NPR \cite{npr} is to detect the forgery fingerprints generated by the up-sampling operation.
In Protocol-4, NPR \cite{npr} still achieved the highest performance results, which demonstrates the effectiveness of adopting the architecture trace.
However, their performance still has a large performance gap compared with the CLIP and DINO series model.
This is mainly caused by the model size and the lack of pre-training knowledge.

\subsection{Out-of-Domain Retrieval Results (Pro-4)}

\begin{table*}[h]
    \centering
    \caption{ACC, mAP \& Recall on Protocol-2 \& 3 in terms of backbones and attribution methods.$r$ denotes the rank used in LoRA and \dag ~refers to CLIP-ViT without pretraining on large-scale data.}
    \label{tab:backbone on Pro2&3}
    \normalsize
    \renewcommand{\arraystretch}{1.0}
    \resizebox{1.0\textwidth}{!}{
    \begin{tabular}{l|ccccc|c|ccccc|c}
    \toprule
    \multirow{4}{*}{\textbf{Methods}} & \multicolumn{6}{c|}{\textbf{Protocol-2}}& \multicolumn{6}{c}{\textbf{Protocol-3}} \\
    \cmidrule{2-13}
    &\multirow{2}{*}{\textbf{Acc}}&\multirow{2}{*}{\textbf{mAP}}&\multirow{2}{*}{\textbf{Ret-mAP}}&\textbf{Ret-Recall}&\textbf{Ret-Recall}&\multirow{2}{*}{\textbf{Avg}}
    &\multirow{2}{*}{\textbf{Acc}}&\multirow{2}{*}{\textbf{mAP}}&\multirow{2}{*}{\textbf{Ret-mAP}}&\textbf{Ret-Recall}&\textbf{Ret-Recall}&\multirow{2}{*}{\textbf{Avg}}\\
    & & & &\textbf{@1\%} &\textbf{@0.1\%} & & & & &\textbf{@1\%} &\textbf{@0.1\%} & \\
    \midrule
    \multicolumn{13}{c}{\textbf{Visual BackBones}}\\
    \midrule
    Xception \cite{xception}
    &0.8899 & 0.9176 & \underline{0.9149} & 0.9515 & 0.8762 & 0.910 
    & 0.7997 & 0.8501 & 0.8519 & 0.8316 & 0.6824 & 0.8031 \\
    F3Net \cite{F3Net}
    &0.8884 & 0.9130 & 0.9106 & 0.9525 & 0.8904 & 0.9110 
    & \underline{0.8220} & 0.8606 & 0.8395 & 0.8433 & 0.7177 & 0.8166 \\
    ResNet-34 \cite{resnet}
    &0.8755 & 0.9044 & 0.8899 & 0.9426 & 0.8923 & 0.9009 
    & 0.7446 & 0.8116 & 0.8158 & 0.8170 & 0.6140 & 0.7606 \\
    Effort \cite{Effort}
    &0.8848 & 0.9154 & 0.7909 & 0.9327 & 0.8943 & 0.8836 
    & 0.7595 & 0.8302 & 0.7337 & 0.9002 & 0.8199 & 0.8087 \\
    CLIP-ViT-FFT \cite{CLIP}
    &0.9021 & 0.9334 & 0.9128 & \textbf{0.9786} & \underline{0.9339} & 0.9322 
    & 0.8091 & 0.8711 & \underline{0.8776} & \underline{0.9294} & \underline{0.8597} & \underline{0.8694} \\
 
    CLIP-ViT-LORA \cite{lora}($r$=1)
    &0.9091 & 0.9459 & 0.8548 & 0.9555 & 0.9254 & 0.9181 
    & 0.8147 & 0.8825 & 0.8029 & \textbf{0.9367} & \textbf{0.8786} & 0.8631 \\
    CLIP-ViT-LORA($r$=64)
    &0.8947 & 0.9296 & 0.7744 & 0.9182 & 0.8785 & 0.8791 
    & 0.7715 & 0.8520 & 0.7334 & 0.8837 & 0.7765 & 0.8034 \\
    DINOv2-L-FFT \cite{dinov2}
    &\textbf{0.9159} & \textbf{0.9511} & 0.9052 & 0.9631 & 0.9329 &\underline{ 0.9337} 
    & \textbf{0.8355} & \textbf{0.9008} & 0.8654 & 0.9259 & 0.8552 & \textbf{0.8766} \\
    DINOv3-L-FFT \cite{dinov3}
    &\underline{0.9130} & \underline{0.9467} & \textbf{0.9245} & \underline{0.9768} & \textbf{0.9507} & \textbf{0.9423} 
    & 0.8198 & \underline{0.8826} & \textbf{0.8818} & 0.9075     & 0.8365 & 0.8656 \\
    \midrule
    \multicolumn{13}{c}{\textbf{Attribution Methods}}\\
    \midrule
    DCT \cite{dct}
    &0.5306 & 0.5561 & 0.3710 & 0.6288 & 0.5343 & 0.5242 
    & 0.5211 & 0.4964 & 0.4124 & 0.4564 & 0.3598 & 0.4492 \\
    DNA-DET \cite{deepfake}
    &0.6235 & 0.5921 & 0.7391 & 0.8176 & 0.7277 & 0.7000 
    & 0.4877 & 0.5337 & 0.6295 & 0.6277 & 0.4223 & 0.5402 \\
    Universal \cite{universal}
    &0.7996 & 0.8456 & 0.5367 & 0.5801 & 0.4945 & 0.6513 
    & 0.6441 & 0.6955 & 0.4847 & 0.4811 & 0.3451 & 0.5301 \\
    LoRAX \cite{lorax}
    &0.8621 & 0.8916 & 0.8173 & 0.8792 & 0.7924 & 0.8485 
    & 0.7361 & 0.8004 & 0.7730 & 0.8234 & 0.6550 & 0.7576 \\
    OCC-CLIP \cite{occ}
    &0.8774 & 0.8952 & 0.8871 & 0.9366 & 0.8894 & 0.8971 
    & 0.7319 & 0.7820 & 0.7690 & 0.7661 & 0.5666 & 0.7231 \\
    NPR \cite{npr}
    &\underline{0.8727} & \underline{0.9028} & \underline{0.8884} & \underline{0.9578} & \textbf{0.9058} & \underline{0.9055} 
    & \underline{0.7597} & \underline{0.8171} & \underline{0.8086} & \underline{0.7975} & \underline{0.6120} & \underline{0.7590} \\
    POSE \cite{progressive}
    &0.8453 & 0.8782 & 0.7713 & 0.8320 & 0.7319 & 0.8117 
    & 0.6977 & 0.7469 & 0.6593 & 0.7475 & 0.5540 & 0.6811 \\
    HIFI-IFDL \cite{hifi}
    &0.8260 & 0.8661 & 0.8588 & 0.9359 & 0.8567 & 0.8687 
    & 0.7422 & 0.7883 & 0.8025 & 0.7842 & 0.6020 & 0.7438 \\
    Repmix \cite{repmix}
    &\textbf{0.8904} & \textbf{0.9173} & \textbf{0.8993} & \textbf{0.9603} & \underline{0.8914} & \textbf{0.9117} 
    & \textbf{0.7655} & \textbf{0.8436} & \textbf{0.8365} & \textbf{0.8317} & \textbf{0.6561} & \textbf{0.7867} \\
    GAN Attribution \cite{attributing}
    &0.7548 & 0.8017 & 0.6221 & 0.6772 & 0.5585 & 0.6829 
    & 0.5784 & 0.6826 & 0.5484 & 0.5098 & 0.3145 & 0.5267 \\
    \bottomrule
    \end{tabular}
    }
\end{table*}

\textbf{Finding-6}. \textbf{CLIP generalizes while DINOv2 collapses in OOD settings.}
Tab.~\ref{tab:backbone on Pro4} shows the performance of each model under Protocol-4 settings, \textit{i.e.}, the retrieval performance under the \textit{Cross Domain, Cross Algorithm} setting. 
Surprisingly, although the DINO models achieve high overall performance in Protocol-2\&3, it lags far behind the CLIP models in Protocol-4.
This shows that the DINO model has a serious over-fitting problem, and the generalization boundary is extremely sharp.
However, the fine-tuned CLIP variants (full fine-tuned, LoRA, and SVD fine-tuned) consistently outperform DINO models by 20-25\% across all metrics on Protocol-4. 
This phenomenon highlights a fundamental architectural difference: the dense local representations ability of DINO tends to overfit specific low-level fingerprints, whereas CLIP's large-scale contrastive pre-training knowledge provides generalized semantic priors that prevent collapse on unseen generative methods.


\textbf{Finding-7}. \textbf{Frequency and structural traces generalize better than specific spatial features.} 
Tab.~\ref{tab:backbone on Pro2&3} evaluates the attribution and retrieval performance of common visual backbones \cite{xception, F3Net, resnet, Effort, CLIP, dinov2, dinov3} and previous attribution methods \cite{dct, DNA-Det, universal, lorax, occ, npr, progressive, hifi, repmix, attributing}.
Surprisingly, the performance of these carefully designed attribution models is inferior to basic ViT models, especially in the retrieval task.
We also find that models that relying on low-level fingerprints can achieve better generalization performance.
For example, although DCT \cite{dct} performed worst on seen domains (Protocol-2\&3), it outperforms more than half of the baselines on Protocol-4. 
This divergence in performance shows that capturing different frequency-domain features through discrete cosine transform provides more advantages in OOD retrieval scenarios than relying on specific spatial features.
The excellent performance of NPR \cite{npr} in Protocol-4 confirms this finding. 
By amplifying the high-frequency fingerprints introduced by the up-sampling operation, NPR \cite{npr} achieves the best performance in Protocol-4.

\subsection{How to Improve Poor OOD Performance?}

To explore effective strategies for alleviating poor OOD generalization performance, we conduct a series of experiments based on CLIP variants.

\begin{table*}[h]
    \centering
    \caption{mAP \& Recall of CLIP variants on Protocol-2\&3.
    }
    \label{tab:clips on Pro2&3}
    \normalsize
    \renewcommand{\arraystretch}{1.0}
    \resizebox{1.0\textwidth}{!}{
    \begin{threeparttable}

    \begin{tabular}{l l|ccccc|c|ccccc|c}
    \toprule
    \multirow{4}{*}{\textbf{Methods}} &\multirow{4}{*}{\textbf{Process}} & \multicolumn{6}{c|}{\textbf{Protocol-2}}& \multicolumn{6}{c}{\textbf{Protocol-3}} \\
    \cmidrule{3-14}
    &&\multirow{2}{*}{\textbf{Acc}}&\multirow{2}{*}{\textbf{mAP}}&\multirow{2}{*}{\textbf{Ret-mAP}}&\textbf{Ret-Recall}&\textbf{Ret-Recall}&\multirow{2}{*}{\textbf{Avg}}
    &\multirow{2}{*}{\textbf{Acc}}&\multirow{2}{*}{\textbf{mAP}}&\multirow{2}{*}{\textbf{Ret-mAP}}&\textbf{Ret-Recall}&\textbf{Ret-Recall}&\multirow{2}{*}{\textbf{Avg}}\\
    & & && &\textbf{@1\%} &\textbf{@0.1\%} & & & & &\textbf{@1\%} &\textbf{@0.1\%} & \\
    \midrule
    CLIP-ViT&- &0.9021 & 0.9334 & \underline{0.9128} & \textbf{0.9786} & 0.9339 & \underline{0.9322} & \underline{0.8091} & \textbf{0.8711} &\underline{0.8776} & 0.9294 & 0.8597 & \textbf{0.8694} \\
    Vanilla-ViT&- 	&0.8667&	0.9038&	0.8749&	0.9290&	0.8254	&0.8799 & 0.7367&	0.8099&	0.8163	&0.8087	&0.6454	&0.7634\\
    \midrule
    CLIP-ViT&\textit{w.} CL&\underline{0.9023} & 0.9320 & \textbf{0.9191} & 0.9703 & 0.9220 & 0.9291 &\textbf{ 0.8193} & \underline{0.8674} &\textbf{ 0.8800} & 0.8965 & 0.7690 & 0.8464 \\
    \midrule
   CLIP-ViT& \textit{w.}LSDA\cite{LSDA}&0.8975 & \underline{0.9377} & 0.9107 & 0.9748 & 0.9391 & 0.9320 & 0.8048 & 0.8587 & 0.8674 & 0.9213 & 0.8144 & 0.8533 \\
    CLIP-ViT&\textit{w.} aug&\textbf{0.9128} & \textbf{0.9449} & 0.9091 & \underline{0.9756} & \textbf{0.9535} & \textbf{0.9392} & 0.7912 & 0.8660 & 0.8643 & 0.9154 & 0.8607 & 0.8595 \\
    Effort\cite{Effort}&\textit{w.} aug&0.8823 & 0.9096 & 0.7730 & 0.9204 & 0.8768 & 0.8724 & 0.7539 & 0.8214 & 0.7282 & 0.9038 & 0.8165 & 0.8048 \\
    \midrule
    CLIP-ViT&\textit{w.} ps=14&0.8783 & 0.9170 & 0.8894 & 0.9622 & 0.9217 & 0.9137 & 0.7816 & 0.8381 & 0.8314 & 0.8831 & 0.7908 & 0.8250 \\
    CLIP-ViT&\textit{w.} ps=7&0.7902 & 0.8385 & 0.8141 & 0.8961 & 0.8034 & 0.8284 & 0.6136 & 0.6805 & 0.6868 & 0.7509 & 0.6005 & 0.6665 \\
    Effort&\textit{w.} ps=14&0.8105&0.8646&	0.6914&	0.8191	&0.7655&	0.7902&	0.6782&	0.7455&	0.6774&	0.8278&	0.6865&	0.7231\\
    \bottomrule
    \end{tabular}

    \end{threeparttable}
    }
\end{table*}
\begin{table}[htbp]
    \centering
        \begin{minipage}[c]{0.48\textwidth}
            \centering
   
            \scriptsize
            \begin{threeparttable}
            \setlength{\tabcolsep}{1.5pt}
            \caption{\small mAP \& Recall of CLIP variants on Protocol-4. 
            CL denotes the contrastive learning and $ps$ indicates the patch size in patch shuffle.
            }
            \label{tab:clips on Pro4}
            \begin{tabular}{l l|ccc|c}
            \toprule
            \multirow{2}{*}{\textbf{Methods}}& \multirow{2}{*}{\textbf{Process}}
            &\multirow{2}{*}{\textbf{Ret-mAP}} &\textbf{Ret-Recall} &\textbf{Ret-Recall}&\multirow{2}{*}{\textbf{Avg}}\\
            & &&\textbf{@1\%}&\textbf{@0.1\%}&\\
            \midrule
            CLIP-ViT & - &\underline{0.5507} & \underline{0.8355} & \underline{0.7358} &\underline{0.7073}\\
            CLIP-ViT & \textit{w/o} FFT&0.2758	&0.6220&	0.4892&0.4623\\
            Vanilla-ViT & - & 0.2553	&0.4050	&0.2199	&0.2934\\
            \midrule
            CLIP-ViT &\textit{w.} CL&0.4233 & 0.7692 & 0.6458 &0.6127\\
            \midrule
            CLIP-ViT &\textit{w.} LSDA\cite{LSDA}&0.4306 & 0.7162 & 0.6070 &0.5846\\
            CLIP-ViT &\textit{w.} aug&0.3672 & 0.6676 & 0.5332& 0.5226\\
            Effort \cite{Effort}&\textit{w.}  aug &\textbf{0.5849} & \textbf{0.8649} & \textbf{0.7915} &\textbf{0.7471}\\
            \midrule
            CLIP-ViT &\textit{w.} ps=14&0.5428 & 0.8196 & 0.7456&0.7026 \\
            CLIP-ViT &\textit{w.} ps=7&0.4349 & 0.7390 & 0.6459& 0.6066\\
            Effort &    \textit{w.}   ps=14&0.4234&	0.6059&	0.4582&	0.4958\\
            \bottomrule
            \end{tabular}
            \end{threeparttable}
            
        \end{minipage}
        \hfill
        \begin{minipage}[c]{0.48\textwidth}

            \scriptsize
            \begin{threeparttable}
            
            \renewcommand{\arraystretch}{1.11}
            \setlength{\tabcolsep}{1.5pt}
            \caption{\small mAP \& Recall of combining features of two methods on Protocol-4. 
            $\dag$ represents the incorporation of Patch Shuffle. By default, patch size is set to $14$ to align with the ViT's token size. $^\ddag$ indicates the use of augmented input data.
            }
            \label{tab:ensembles on Pro4}
            \begin{tabular}{l l|ccc|c}
            \toprule
            \multirow{2}{*}{\textbf{Method1}}& \multirow{2}{*}{\textbf{Method2}}
            &\multirow{2}{*}{\textbf{Ret-mAP}} &\textbf{Ret-Recall} &\textbf{Ret-Recall}&\multirow{2}{*}
            {\textbf{Avg}}\\
            \setlength{\tabcolsep}{1.5pt}
            & &&\textbf{@1\%}&\textbf{@0.1\%}&\\
            \midrule
            CLIP-ViT& - &0.5507&0.8355&0.7358&0.7073\\
            \midrule
            CLIP-ViT$^\dag$&CLIP-ViT&0.6138 & 0.8651 & 0.8019& 0.7602\\
            CLIP-ViT&Effort$^\ddag$&\underline{0.6294} &\textbf{ 0.9070} & \underline{0.8469} &\underline{0.7944}\\
            CLIP-ViT$^\dag$&Effort$^\ddag$&\textbf{0.6536} & \underline{0.9056} & \textbf{0.8598}&\textbf{0.8063}\\
            \midrule
            DINOv2&Effort$^\ddag$&0.5248	&0.8101	&0.6946&	0.6765\\
            DINOv2&CLIP-ViT&0.5034	&0.8078&	0.7186&	0.6766\\
            DINOv2&CLIP-ViT$^\dag$&0.5263	&0.8329	&0.7639	&0.7077\\
            \bottomrule
            \end{tabular}
            \end{threeparttable}
            
        \end{minipage}
\end{table}

\begin{table}[t]
    \centering
        \begin{minipage}[c]{0.49\textwidth} \centering
        \includegraphics[width=0.95\linewidth]{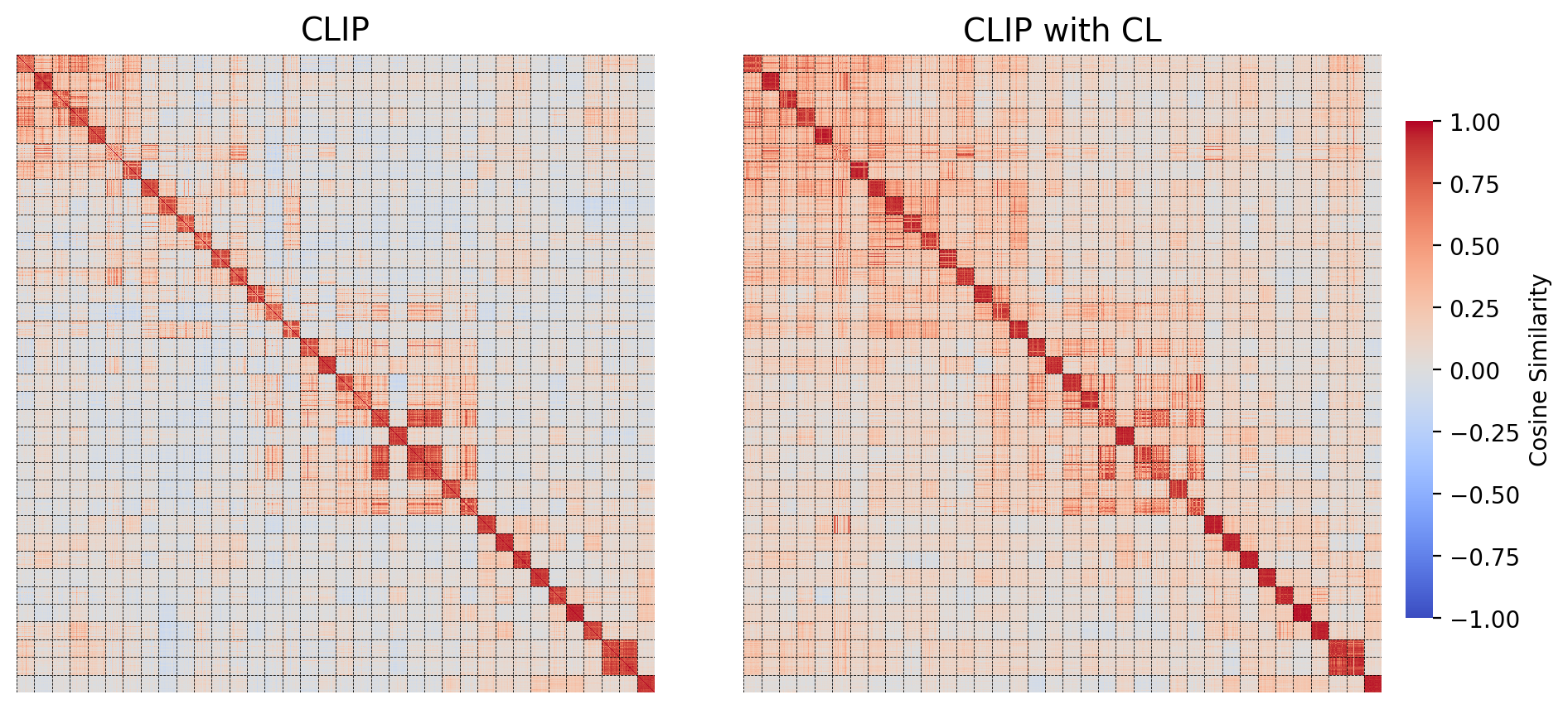}
        \captionof{figure}{\small Cosine similarity heatmaps of features on Protocol-2. CL indicates contrastive learning. A redder color indicates a higher similarity.}
        \label{fig:heatmap}

        \end{minipage}
        \hfill
        \begin{minipage}[c]{0.49\textwidth}
        \includegraphics[width=0.95\linewidth]{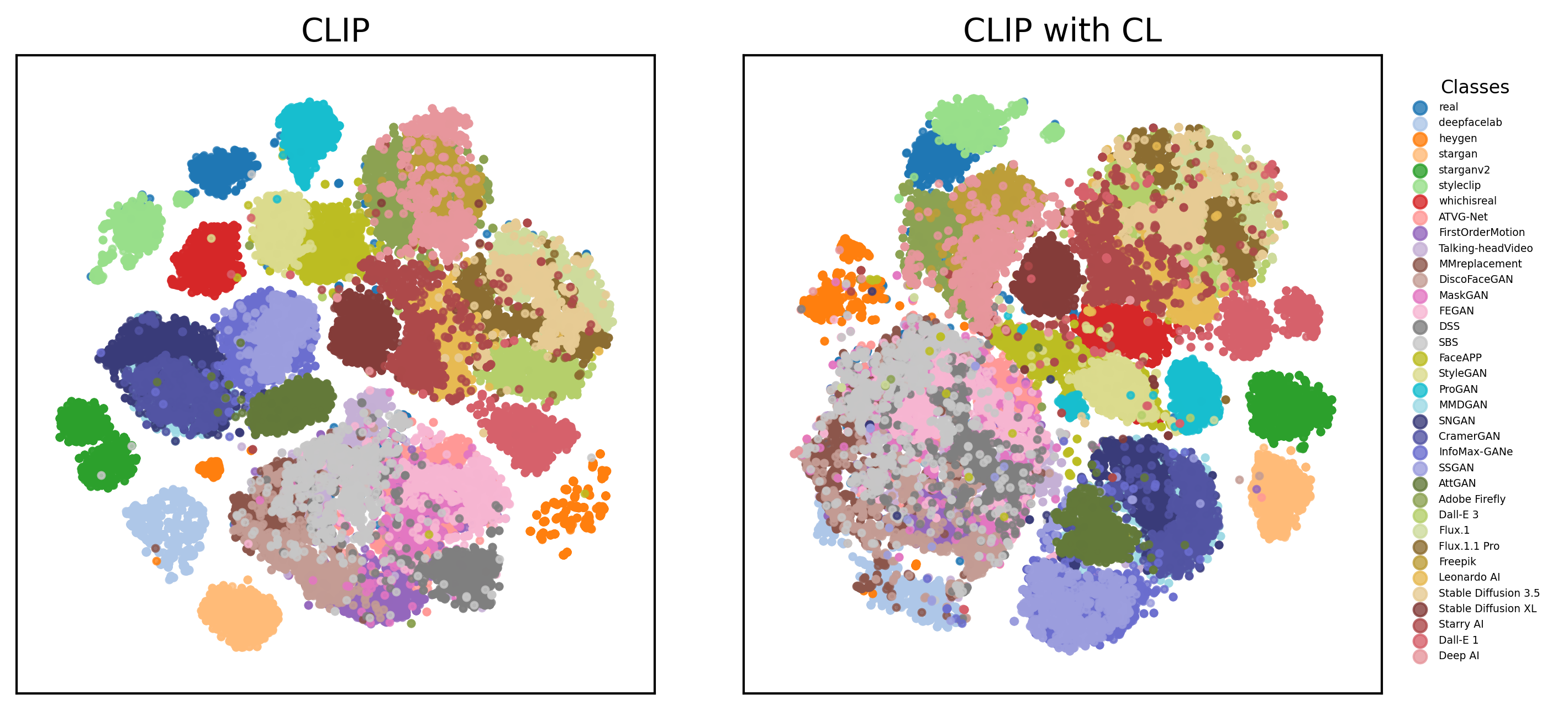}
        \captionof{figure}{\small t-SNE visualization of features on 36 out-of-domain algorithms on Protocol 4. CL indicates contrastive learning.}
        \label{fig:tsne}
        \end{minipage}
\end{table}
\textbf{Finding-8}. 
\textbf{Contrastive learning biases the feature space and hurts OOD generalization performance.}
Traditional attribution methods \cite{DNA-Det, li2024boosting} usually adopt contrastive learning (CL) to enhance the feature representation ability, while the counterintuitive phenomenon in Tab. \ref{tab:clips on Pro4} can not support this experience.
Compared with the naive full fine-tune model, CL brings nearly 9\% performance drop.
To explain this phenomenon, we carried out a series of explorations and found that this performance is caused by \textit{Overfitting} and \textit{Semantic Space Collapse} during contrastive learning. 
Fig.~\ref{fig:heatmap} visualizes the pairwise cosine similarity heatmaps between feature representations in Protocol-2 (in-domain evaluation).
The diagonal represents intra-class similarity and other regions represent inter-class similarity.
It can be found that the model with CL has higher similarity between different classes, which implies higher confusion between known classes.
Consequently, this confusion slightly degrades the in-domain performance, resulting 0.9291 average performance on Protocol-2 (CLIP-ViT achieves 0.9322), as show in Tab.~\ref{tab:clips on Pro2&3}.
In addition, we adopted t-SNE diagram to visually compare the features before and after using comparative learning on Protocol-4 in OOD setting.
The degradation of the feature separability confirms the aforementioned analysis.
Due to length constraints, more detailed and comprehensive analysis is listed in the Appendix \ref{app:cl_feature_space_analysis}.

\textbf{Findings-9. The OOD generalization ability of CLIP-ViTs relies heavily on pre-trained semantic priors.}
We compared three ViT model variants to assess the importance of pre-trained knowledge: a \textit{Vanilla ViT} \cite{ViT} without contrastive language-image pre-training, a standard \textit{CLIP-ViT with full fine-tune}, and a \textit{Frozen CLIP-ViT} (zero-shot setting). 
As shown in Tab.~\ref{tab:clips on Pro2&3}, the fine-tuned vanilla ViT achieves respectable 0.8799 and 0.7634 average performance on Protocol-2\&3, respectively. This performance lags behind the fine-tuned CLIP-ViT with average performance 0.9322 and 0.8694. 
However, the fine-tuned vanilla ViT only achieve 0.2934 on Protocol-4, which is extremely lower than the fine-tuned CLIP-ViT with 0.7073, as shown in Tab. \ref{tab:clips on Pro4}. 
Even without any fine-tuning, the zero-shot CLIP-ViT achieves 0.4623 on Protocol-4.
These evidences directly confirm that the attribution robustness of CLIP-ViT models benefits from the large-scale pre-trained priors.

\textbf{Finding-10}. \textbf{Low-level and high-level features are complementary for OOD generalization.}
We conducted a series of experiments to study the complementarity between \textit{Low-level Fingerprints} and \textit{High-level Semantics}, as shown in Tab.~\ref{tab:ensembles on Pro4}.
To separate these features, we adopt two specialized models: Effort \cite{Effort} with data augmentations that emphasizes high-level semantics, while a patch-shuffled CLIP-ViT \cite{CLIP} is forced to focus on only low-level fingerprints (due to space limitations, more details are provided in Appendix~\ref{finding11}).
Individually, the standard CLIP-ViT achieves an average performance of 70.73\% on Protocol-4, whereas the patch-shuffled CLIP-ViT shows an average performance drop of about 2\% across Protocols.
Firstly, combining the standard CLIP-ViT with models focusing on either high-level semantics (Effort) or low-level fingerprints (patch-shuffled CLIP-ViT) both bring performance improvements.
Specifically, fusing the standard CLIP-ViT  with the patch-shuffled CLIP-ViT brings a 6\% improvement on Protocol-4, while fusing it with Effort gains a 9\% improvement.
Secondly, the combination of specialized high-level and low-level models achieves the best results.
Combining the patch-shuffled CLIP-ViT with the Effort achieves a remarkable 10\% boost compared to a single standard CLIP-ViT.
Interestingly, combining Effort with the stronger standard CLIP-ViT performs worse than pairing it with the weaker patch-shuffled variant.
We attribute this to feature redundancy and domain bias: standard CLIP-ViT contains mixed features with domain-specific bias, which interferes with Effort’s robust semantics, whereas the patch-shuffled model provides orthogonal low-level cues.
These results suggest that effective OOD generalization is not achieved by naively stacking strong models (see Appendix~\ref{app:finding10} for further analysis).
Pairing pure high-level semantics with pure low-level fingerprints forms a synergistic representation, enabling more robust and accurate source attribution.

\section{Limitation and Discussion}
\textbf{The semantics of the collected data are not strictly aligned}.
Because the entire dataset largely refers to the existing deepfake datasets, and only 8 algorithms (strictly semantic alignment on CelebA) have been synthesized. 
Therefore, the performance metrics may not reflect the true attribution performance. The retrieval \textit{mAP} and \textit{Recall} may be affected by semantic information.
A straightforward solution is to re-synthesize all algorithms based on fixed real semantic data. However, this would incur an enormous workload. We leave this challenging task for future work.

\textbf{No novel framework are proposed in this paper}.
This paper presents a systematic evaluation of deepfake detection and attribution tasks, and explores how classical strategies (\textit{e.g.}, patch shuffle and data augmentation) can be adopted to explore the performance upper bound of existing models. 
Through model combination, we further conduct a complementary analysis of high-level semantic and low-level fingerprint features.
Due to space and time limitations, we did not propose novel mechanisms and frameworks for deepfake attribution task.
Based on our empirical findings, we highlight several promising research directions, including \textit{Latent Space Feature Augmentation}, the \textit{Disentanglement of High‑level Semantic and Low‑level Fingerprint}, and \textit{Advanced Data Augmentation Strategies}. 
We believe this work can provide valuable insights and serve as a solid foundation for future research on deepfake attribution architecture exploring.

\textbf{A temporal gap exists between DFG and the latest generative models}.
This work was initiated in 2025, during which we collected a series of SOTA open-source and closed-source algorithms. 
Representative examples include open-source methods such as \textit{HunyuanImage-3.0} \cite{HunyuanImage-3.0} and \textit{Qwen-Image} \cite{Qwen-Image}, as well as closed-source solutions including \textit{Nano Banana} \cite{nanobanana} and \textit{GPT-4o} \cite{gpt-4o}.
Over the past year, the iteration speed of AIGC algorithms has far exceeded general expectations.
Since adjusting the algorithm composition of the DFG benchmark is rather cumbersome, we retain the original 100 algorithms and do not supplement the most recently released generative models.
Nevertheless, this limitation does not undermine the validity and generalization of the core findings in this paper. 

\section{Conclusions and Limitations}
This work presents a comprehensive DFG benchmark for deepfake detection and attribution. 
The DFG covers 100 facial forgery methods and 2M images with a much larger scale than prior DFA benchmarks. 
Additionally, 4 practical evaluation protocols are established, including a novel retrieval-based attribution paradigm that better fits active defense scenarios such as blacklist registration. 
Extensive experiments with over 2,000+ evaluations are conducted to summarize 10 insightful findings. 
DFG provides a unified benchmark for systematically evaluating deepfake attribution methods and guiding the development of robust and generalizable attribution systems.


{
\small
\bibliographystyle{IEEEtran}
\bibliography{reference}
}

\appendix
\clearpage
\section{Detailed Information of the DFG Benchmark}
\textbf{Detailed list of 100 algorithms}.
Tab.~\ref{tab:DFG-100-algos} provides a detailed enumeration of the 100 algorithms in the DFG benchmark, along with the corresponding domains.
Since the first 92 algorithms are collected from existing Deepfake and DFA datasets \cite{FF++, DF40, Forgerynet, DFFD, DNA-Det, OSMA, wild, DiFF, Diffusionface}, the domain is relatively rich.
In the \textit{Domain} column, \textit{Multi-1} represents CREMA-D, VoxCeleb2, RAVDESS, and AVSpeech. \textit{Multi-2} represents FFHQ, CelebA and FF++. \textit{Multi-3} represents VoxCeleb2 and CelebA.

\textbf{Detailed data partition in 4 Protocols}.
Tab.~\ref{tab:pro-1-traintest}-\ref{tab:pro-4-traintest} provide a detailed display of the algorithm settings under 4 evaluation protocols.
The DFG benchmark not only enables extended evaluation in traditional Deepfake tasks but also provides rich settings on DFA evaluations.
The Protocol-2 and Protocol-3 focus on \textit{Same Domain, Same Algorithm} and \textit{Cross Domain, Same Algorithm}, respectively.
Protocol-4 focuses on novel forgery attacks in \textit{Cross Domain, Cross Algorithm} setting, which is most similar to threats in the real world.

\begin{figure*}[h]
    \centering
    \includegraphics[width=1.0\linewidth]{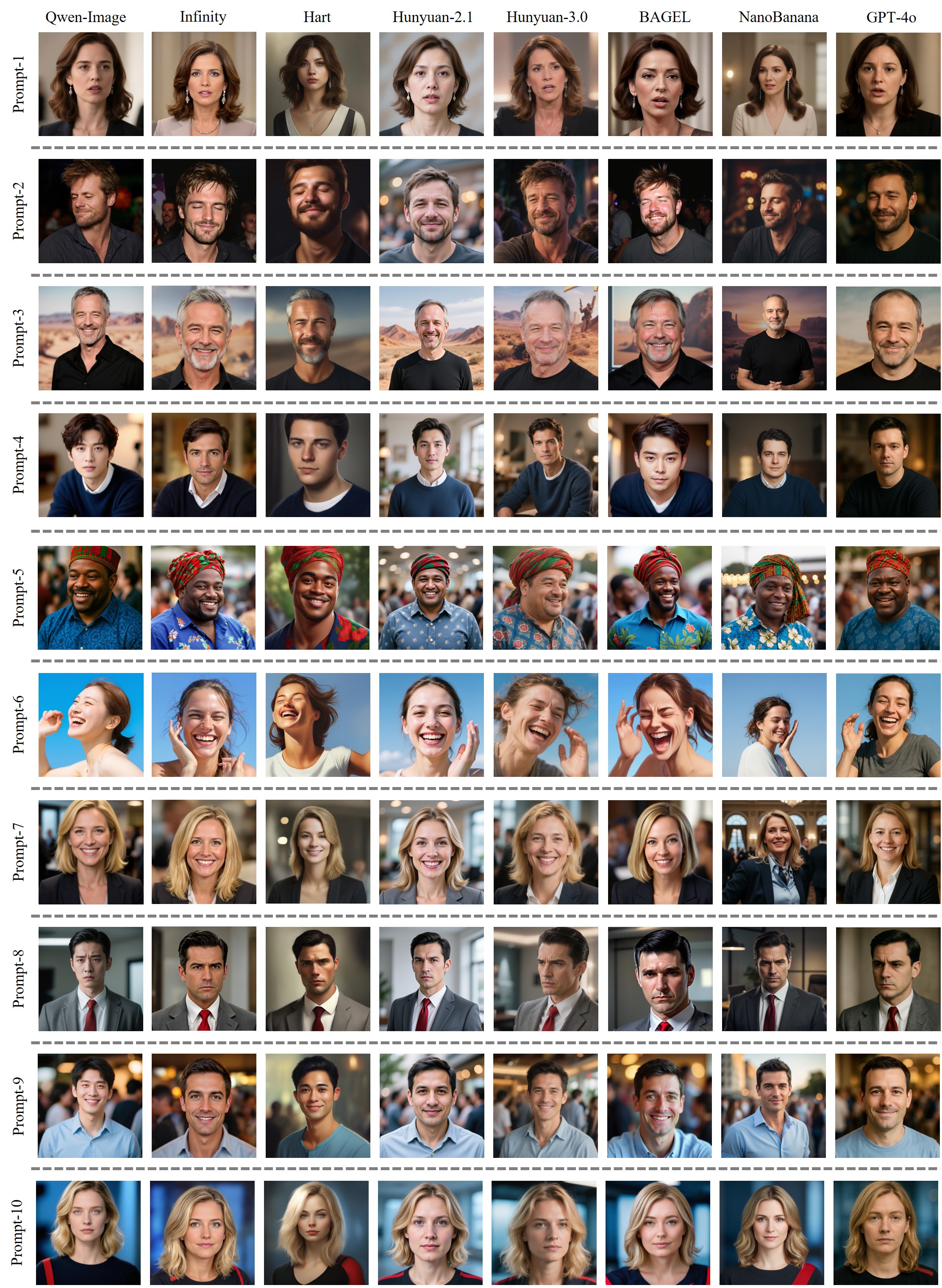}
    \caption{Visualization of 8 synthesized algorithms.}
    \label{fig:8_Algo_Images}
\end{figure*}

\textbf{Visualization of 8 Synthesized Algorithms}.
We adopt real facial images from the CelebA and leverage the Qwen3-VL \cite{Qwen3-VL} to generate corresponding textual descriptions.
In Fig.~\ref{fig:8_Algo_Images}, we present 10 representative cases for intuitive illustration, with the detailed prompts listed in Tab.~\ref{tab:prompt_list}.
It can be observed that NanoBanana \cite{nanobanana} and GPT-4o \cite{gpt-4o} achieve an extremely high level of visual realism.
Meanwhile, open-source algorithms including Qwen-Image \cite{Qwen-Image}, Infinity \cite{Infinity}, and HunyuanImage-3.0 \cite{HunyuanImage-3.0} are also capable of producing highly realistic human portraits.
By contrast, the visual quality of Hart \cite{Hart}, HunyuanImage-2.1 \cite{HunyuanImage-2.1} and Bagel \cite{BAGEL} is relatively inferior.

\section{Detailed Results in Protocol-1}
Tab.~\ref{tab:pro1-types} and Tab.~\ref{tab:pro1-architectures} in the main text categorize and summarize the results of the deepfake methods in Protocol-1.
We list all detailed results in Tab.~\ref{tab:acc-all} and Tab.~\ref{tab:auc-all}.
It should be noted that in Protocol-1, only 59 new algorithms were evaluated, excluding FF++ \cite{FF++} and DF40 \cite{DF40}.
This design is mainly based on two considerations: the algorithms adopted in DF40 are no longer innovative, and this dataset has been widely adopted for evaluating the generalization ability of deepfake models.
To avoid duplicate evaluation efforts, we exclude all DF-40 algorithms in Protocol-1.

\section{More Details of Finding-8}
\label{app:cl_feature_space_analysis}

\begin{figure}[h]
	\centering
	\begin{subfigure}{0.24\linewidth}
		\centering
		\includegraphics[width=1\linewidth]{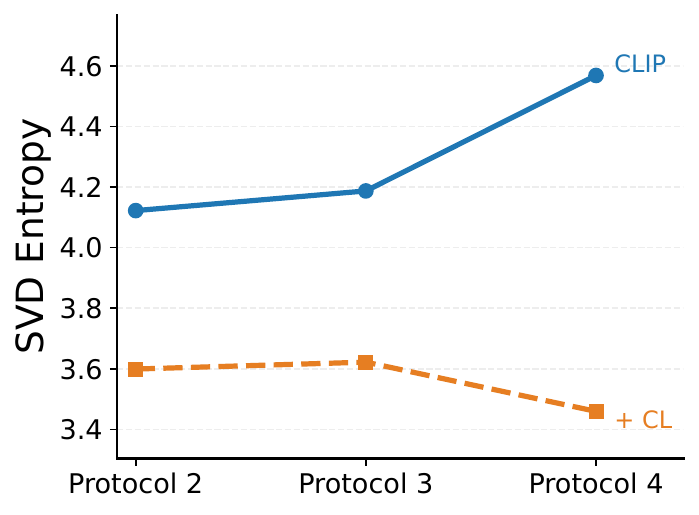}
		\caption{SVD\_Entropy}
		\label{ma}
	\end{subfigure}
	\centering
	\begin{subfigure}{0.24\linewidth}
		\centering
		\includegraphics[width=1\linewidth]{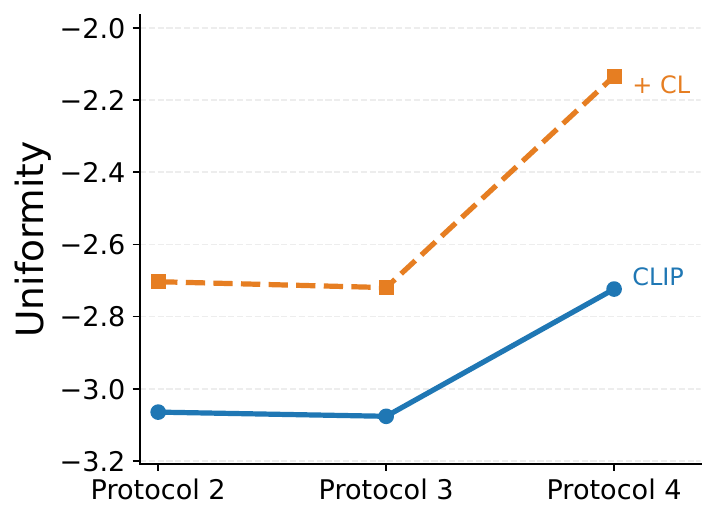}
		\caption{Uniformity~\cite{uniformity}}
		\label{mb}
	\end{subfigure}
    \centering
	\begin{subfigure}{0.24\linewidth}
		\centering
		\includegraphics[width=1\linewidth]{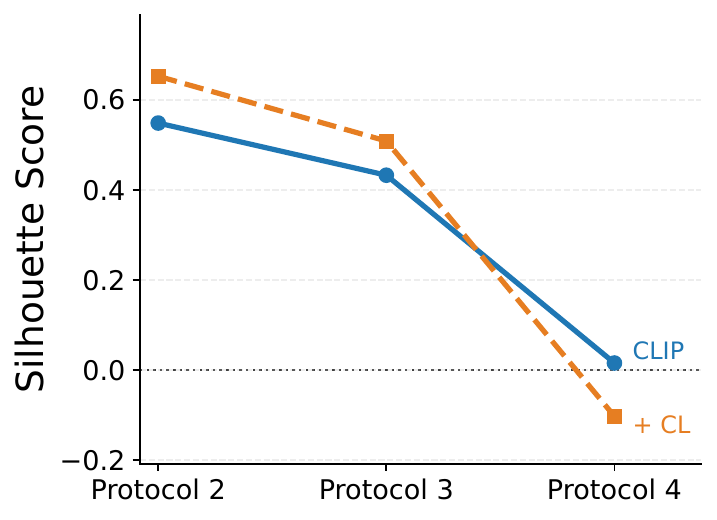}
		\caption{Silhouette\_Score~\cite{silhouettes}}
		\label{mc}
	\end{subfigure}
    \begin{subfigure}{0.24\linewidth}
		\centering
		\includegraphics[width=1\linewidth]{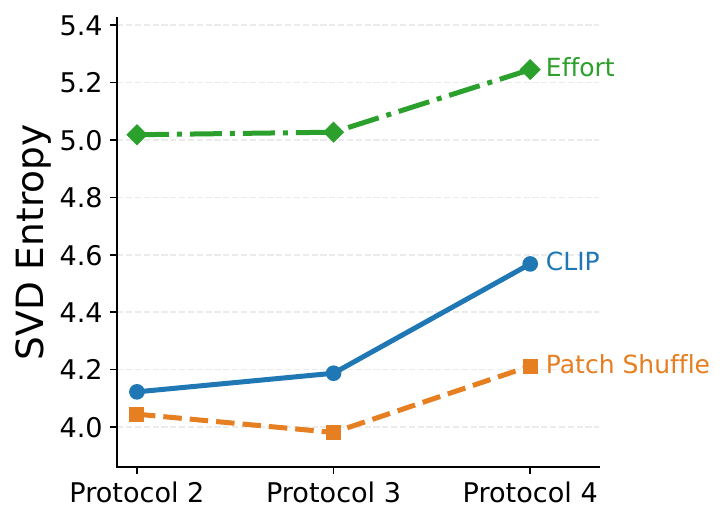}
		\caption{SVD\_Entropy}
		\label{md}
	\end{subfigure}
	\caption{\textbf{Comparison of feature space properties across evaluation protocols.} (a-c) SVD Entropy, Uniformity, and Silhouette Score for the CLIP-ViT and CLIP-ViT equipped with contrastive learning. (d) SVD Entropy comparison among Effort, CLIP, and CLIP-ViT equipped with patch shuffle.}
	\label{fig:metrics}
\end{figure}

\subsection{Analysis}
To further verify the feature-space degradation discussed in Finding-8, we quantitatively evaluate the learned representation space using \textit{Singular Value Decomposition Entropy} (SVD Entropy), \textit{Uniformity}~\cite{uniformity}, and \textit{Silhouette Score}~\cite{silhouettes}, as shown in Fig.~\ref{fig:metrics}.

\textbf{SVD entropy.}
Similar to effective rank~\cite{effecientrank}, SVD entropy measures the information richness and effective dimensionality of the feature space.
A higher SVD entropy indicates that the representation occupies more informative dimensions, while a lower value suggests dimensional collapse.
As shown in Fig.~\ref{ma}, the addition of contrastive learning results in a sharp drop in SVD entropy.
For example, on Protocol 2, the entropy decreases from $4.122$ to $3.599$.
This confirms that contrastive learning leads to a severe loss of informative dimensions and compresses the representation space into a lower-dimensional subspace.

\textbf{Uniformity.}
We further use the uniformity metric~\cite{uniformity} to evaluate how evenly features are distributed on the hypersphere.
A lower uniformity value indicates a more uniformly distributed feature space.
However, as shown in Fig.~\ref{mb}, contrastive learning worsens the uniformity score, increasing it from $-3.06$ to $-2.70$.
This indicates that the feature space becomes less continuous and less evenly distributed.
Instead of spreading samples across the hypersphere, contrastive learning causes features to clump together in restricted and crowded regions.

\textbf{Silhouette score.}
We also compute the silhouette score~\cite{silhouettes}, which measures clustering quality by comparing intra-class compactness and inter-class separation.
A negative silhouette score indicates severe overlap between different classes.
As shown in Fig.~\ref{mc}, although contrastive learning artificially improves the clustering score on seen domains by forcing training classes to be compact, its score drops drastically on unseen domains.
In particular, on Protocol 4, the silhouette score of the contrastive model decreases to $-0.103$, whereas the baseline remains positive at $0.015$.
This result demonstrates that the compactness learned by contrastive learning does not generalize to unseen generative methods.
Instead, it amplifies inter-class confusion in out-of-domain scenarios.

Overall, these quantitative results support the conclusion in Finding-8: contrastive learning overfits the model to the seen training taxonomy, reduces feature diversity, and collapses the continuous semantic space.
This explains why contrastive learning slightly hurts in-domain discrimination but severely damages out-of-domain generalization.

The negative impact of this distorted space becomes more obvious when evaluating the model on unseen domains. 
While the baseline maintains reasonable robustness with scores of 0.8694 on Protocol 3 and 0.7073 on Protocol 4, the contrastive learning model suffers a severe performance drop, yielding only 0.8464 and 0.6127, respectively. 
This decline occurs because the continuous feature space has been disrupted, causing the features of unseen new generative models to scatter unpredictably in the space.
This phenomenon is clearly visualized using t-SNE in Fig.~\ref{fig:tsne}. 
While the baseline (Fig.~\ref{fig:tsne}a) keeps unseen methods relatively separated, the contrastive model (Fig.~\ref{fig:tsne}b) mixes them into a chaotic and overlapping state. 
Additional quantitative analyses using SVD entropy, Uniformity~\cite{uniformity}, and Silhouette score~\cite{silhouettes} are provided in Appendix~\ref{app:cl_feature_space_analysis}, which further confirm the feature-space collapse caused by contrastive learning.

\subsection{Implementation of Uniformity and SVD Entropy}
\label{app:metrics_implementation}

In this section, we provide the implementation details of the two representation-space metrics used in our analysis: uniformity and Singular Value Decomposition (SVD) entropy. 
Given a feature matrix $\mathbf{Z} \in \mathbb{R}^{N \times d}$, where $N$ is the number of samples and $d$ is the feature dimension, all features are first converted to floating-point tensors. For uniformity, features are additionally $\ell_2$-normalized before computing pairwise distances.

\subsubsection{Uniformity}

\begin{algorithm}[H]
\caption{Uniformity Calculation}
\label{alg:uniformity}
\begin{algorithmic}[1]
\Require Feature matrix $\mathbf{Z} \in \mathbb{R}^{N \times d}$, temperature $t=2.0$, maximum sample size $M=20{,}000$
\Ensure Uniformity score $\mathcal{L}_{\mathrm{uniform}}$
\If{$N > M$}
    \State Randomly sample $M$ rows from $\mathbf{Z}$
\EndIf
\State Normalize each feature: $\hat{\mathbf{z}}_i \leftarrow \mathbf{z}_i / \|\mathbf{z}_i\|_2$
\State Compute pairwise squared distances: $d_{ij} \leftarrow \|\hat{\mathbf{z}}_i - \hat{\mathbf{z}}_j\|_2^2$ for all $i < j$
\State Compute $\mathcal{L}_{\mathrm{uniform}} \leftarrow \log \left( \frac{1}{|\mathcal{P}|} \sum_{(i,j)\in \mathcal{P}} \exp(-t d_{ij}) \right)$
\State \Return $\mathcal{L}_{\mathrm{uniform}}$
\end{algorithmic}
\end{algorithm}
Uniformity measures how evenly features are distributed on the unit hypersphere. Following~\cite{uniformity}, we compute the uniformity score as:
\begin{equation}
    \mathcal{L}_{\mathrm{uniform}}(\mathbf{Z})
    =
    \log 
    \mathbb{E}_{i \neq j}
    \left[
    \exp
    \left(
    -t \left\|
    \frac{\mathbf{z}_i}{\|\mathbf{z}_i\|_2}
    -
    \frac{\mathbf{z}_j}{\|\mathbf{z}_j\|_2}
    \right\|_2^2
    \right)
    \right],
    \label{eq:uniformity}
\end{equation}
where $t$ is a temperature hyperparameter and is set to $2.0$ in our experiments. A smaller, i.e., more negative, value indicates that features are more uniformly spread over the hypersphere.
For computational efficiency, when the number of samples exceeds $20{,}000$, we randomly subsample $20{,}000$ features to estimate this metric.

\subsubsection{SVD Entropy}

\begin{algorithm}[h]
\caption{SVD Entropy Calculation}
\label{alg:svd_entropy}
\begin{algorithmic}[1]
\Require Feature matrix $\mathbf{Z} \in \mathbb{R}^{N \times d}$, numerical threshold $\epsilon=10^{-7}$
\Ensure SVD entropy $\mathcal{H}_{\mathrm{SVD}}$
\State Center the features: $\tilde{\mathbf{Z}} \leftarrow \mathbf{Z} - \mathrm{mean}(\mathbf{Z}, \mathrm{dim}=0)$
\State Compute singular values $\{\sigma_i\}_{i=1}^{r}$ of $\tilde{\mathbf{Z}}$
\State Compute spectral energy: $\lambda_i \leftarrow \sigma_i^2$
\State Normalize energy: $p_i \leftarrow \lambda_i / \sum_{j=1}^{r} \lambda_j$
\State Remove numerically unstable entries where $p_i \leq \epsilon$
\State Compute $\mathcal{H}_{\mathrm{SVD}} \leftarrow -\sum_i p_i \log p_i$
\State \Return $\mathcal{H}_{\mathrm{SVD}}$
\end{algorithmic}
\end{algorithm}

SVD entropy measures the effective utilization of feature dimensions. Given the feature matrix $\mathbf{Z}$, we first center it by subtracting the feature mean:
\begin{equation}
    \bar{\mathbf{z}} = \frac{1}{N}\sum_{i=1}^{N}\mathbf{z}_i,
    \qquad
    \tilde{\mathbf{z}}_i = \mathbf{z}_i - \bar{\mathbf{z}}.
\end{equation}
We then compute the singular values of $\tilde{\mathbf{Z}}$:
\begin{equation}
    \sigma_1, \sigma_2, \ldots, \sigma_r = \mathrm{SVD}(\tilde{\mathbf{Z}}),
\end{equation}
where $r = \mathrm{rank}(\tilde{\mathbf{Z}})$. The normalized spectral energy is defined as:
\begin{equation}
    p_i = \frac{\sigma_i^2}{\sum_{j=1}^{r} \sigma_j^2}.
\end{equation}
The SVD entropy is then computed as:
\begin{equation}
    \mathcal{H}_{\mathrm{SVD}}(\mathbf{Z})
    =
    - \sum_{i=1}^{r} p_i \log p_i.
    \label{eq:svd_entropy}
\end{equation}
A higher SVD entropy suggests that the representation uses more informative dimensions, while a lower value indicates that the feature space is concentrated in fewer dominant directions, suggesting potential dimensional collapse.

\section{More Analysis of Finding-10}
\label{app:finding10}
\subsubsubsection{\textbf{More details of the fingding-10}}

A further comparison reveals that combining Effort with the stronger standard CLIP-ViT performs worse than pairing it with the weaker patch-shuffled CLIP-ViT.
We attribute this phenomenon to feature redundancy and domain bias.
The standard CLIP-ViT learns a mixture of low-level and high-level features, where high-level semantics are often biased toward seen domains.
When combined with Effort, these biased features overlap with Effort’s semantic representations and introduce interference.
In contrast, the patch-shuffled CLIP-ViT provides purely low-level signals, as spatial semantics are removed by patch shuffling.
This minimizes feature overlap while maximizing complementarity between models.
Nevertheless, not all model combinations are beneficial.
For example, as shown in Tab.~\ref{tab:ensembles on Pro4}, combining DINOv2 with other models degrades performance, likely due to severe feature interference and redundancy.
These observations further confirm that naive ensembling of strong models is suboptimal for OOD generalization.
Instead, constructing orthogonal feature spaces—by combining pure low-level fingerprints and high-level semantics—is the key to achieving robust cross-domain generalization.

\subsubsubsection{\textbf{SVD Entropy Analysis of Effort}}

To further support the analysis in Finding-10, we measure the Singular Value Decomposition (SVD) entropy of different models, as shown in Fig.~\ref{md}.
SVD entropy serves as a metric for the information richness and effective dimensionality of the feature manifold.
A higher SVD entropy indicates that the representation preserves more informative directions and is less likely to suffer from feature-space collapse.
The results show that Effort maintains consistently higher SVD entropy across all protocols.
This suggests that Effort better preserves the pre-trained CLIP-ViT representation space, while full fine-tuning tends to reduce the diversity and dimensionality of the feature manifold.
This difference becomes especially clear on the fully unseen Protocol 4.
The entropy score of Effort reaches $5.2455$, significantly surpassing the $4.5680$ of the fully fine-tuned CLIP-ViT and the $4.2117$ of its patch-shuffled variant.
This quantitative evidence supports our conclusion that Effort's out-of-domain robustness comes from preserving the pre-trained semantic structure.
Unlike CLIP-ViT-FFT, which aggressively adapts the feature space to seen-domain fingerprints, Effort retains a richer and more continuous representation manifold.
Therefore, when low-level artifacts are unreliable or corrupted, Effort can still rely on robust high-level semantic cues, leading to better generalization on unseen generative methods.

\section{Analysis on Low-level Fingerprints and High-level Semantics}
\label{finding11}
To investigate the differences between the fully fine-tuned CLIP-ViT and Effort, which preserves the vast majority of pre-trained semantic information, we conducted patch shuffling and data augmentation experiments as shown in Tab.~\ref{tab:clips on Pro2&3} and Tab.~\ref{tab:clips on Pro4}.
First, for CLIP-ViT-FFT, when applying patch shuffling at the default $14 \times 14$ resolution, which disrupts global spatial semantics while preserving local details, CLIP-ViT-FFT exhibits remarkable robustness. It shows minor performance drops of $1.85\%$ and $4.44\%$ on Protocols 2 and 3, and a negligible drop of merely $0.47\%$ on the unseen Protocol 4. 
This indicates that destroying spatial semantics barely affects its out-of-domain performance. 
However, reducing the shuffle granularity from $14 \times 14$ to $7 \times 7$ leads to a significant performance decline, as finer shuffling destroys not only global semantics but also the local spatial consistency aligned with the Vision Transformer tokenization scale. 
Conversely, corrupting low-level information yields the exact opposite effect. 
While standard data augmentations and latent space domain augmentation(\textit{i.e.}, LSDA) achieve comparable or even better performance on the seen domains of Protocol- 2\&3, they cause massive performance drops of $18.47\%$ and $12.27\%$ respectively on Protocol-4. 
This clear difference exposes a critical mechanism: full fine-tuning causes the model's high-level representations to severely overfit to the seen domains. 
As a result, the model's generalization ability relies almost entirely on local, low-level fingerprints. 
However, when data augmentation corrupts these low-level details, the model is forced to rely more on its high-level features. 
Since these high-level features are already overfitted, this forced reliance inevitably leads to a drop in robustness and poor generalization.
In contrast, the Effort  exhibits improved OOD generalization when subjected to data augmentations. Unlike CLIP\_L\_FFT, Effort freezes the vast majority of CLIP's pre-trained semantic features, this allows the model to map unseen generative methods into distinct clusters within this broad semantic space and effectively prevents the high-level overfitting observed in CLIP-ViT. When augmentations like blurring corrupt low-level details, Effort is forced to concentrate on its well-preserved, high-level semantic features, which inherently enhances its robustness.

\begin{table*}[htbp]
\centering
\caption{Summary of 100 facial forgery algorithms in DeepfakeGenome benchmark.}


\label{tab:DFG-100-algos}
\tiny
\renewcommand{\arraystretch}{0.65}
\setlength{\tabcolsep}{2pt}
\begin{tabular}{lllllllllll}
\toprule
\textbf{Dataset} & \textbf{Task} & \textbf{ID} & \textbf{Algorithm} & \textbf{Venue} & \textbf{Type} & \textbf{Domain} & \textbf{FF++ Train} & \textbf{FF++ Test} & \textbf{Other Real} & \textbf{Other Fake} \\
\midrule
\multirow{4}{*}[-1.5ex]{FF++ \cite{FF++}}
 & FS     & 001 & Deepfakes \cite{Deepfakes}          & None & GAN     & FF++          & 22928 & 4454  & -           & - \\
 & FS     & 002 & FaceShifter \cite{faceshifter}        & ArXiv 2019 & CNN     & FF++          & 22968 & 4479  & -           & - \\
 & FR   & 003 & NeuralTextures \cite{NeuralTextures}     & TOG 2019 & GAN     & FF++          & 22973 & 4479  & -           & - \\
 & FR   & 004 & Face2Face \cite{Face2face}          & CVPR 2016 & CG      & FF++          & 22991 & 4480  & -           & - \\
\midrule
\multirow{26}{*}[-12ex]{DF40 \cite{DF40}}
 & FS     & 005 & FSGAN \cite{Fsgan}              & ICCV 2019 & GAN     & FF++ \& CDF   & 21815 & 4455  & CDF-5.5k    & 19601 \\
 & FS     & 006 & SimSwap \cite{Simswap}            & MM 2020 & GAN     & FF++ \& CDF   & 31350 & 4468  & CDF-5.5k    & 20322 \\
 & FS     & 007 & BlendFace \cite{Blendface}          & ICCV 2023 & GAN     & FF++ \& CDF   & 22513 & 4467  & CDF-5.5k    & 20323 \\
\bottomrule
\end{tabular}%

\end{table*}

\begin{table}[htbp]
    \centering
    \caption{Protocol-1 training and testing algorithms.}
    \label{tab:pro-1-traintest}
    \tiny
    \begin{tabular}{p{0.25\textwidth} p{0.65\textwidth}}
    \toprule
    \textbf{Training Set} & \textbf{Algorithms} \\
\bottomrule
    \end{tabular}
    \end{table}
\begin{table}[htbp]
    \centering
    \caption{Protocol-2 training and testing algorithms.}
    \label{tab:pro-2-traintest}
    \tiny
    \resizebox{\textwidth}{!}{%
    \begin{tabular}{p{0.45\textwidth} p{0.45\textwidth}}
    \toprule
    \textbf{Training Set} & \textbf{Test Set} \\
\bottomrule
    \end{tabular}%
    }
    \end{table}
\begin{table}[htbp]
    \centering
    \caption{Protocol-3 training and testing algorithms.}
    \label{tab:pro-3-traintest}
    \tiny
    \resizebox{\textwidth}{!}{%
    \begin{tabular}{p{0.45\textwidth} p{0.45\textwidth}}
    \toprule
    \textbf{Training Set} & \textbf{Testing Set} \\
\bottomrule
    \end{tabular}%
    }
    \end{table}
\begin{table}[htbp]
    \centering
    \caption{Protocol-4 training and testing algorithms.}
    \label{tab:pro-4-traintest}
    \tiny
    \begin{tabular}{p{0.25\textwidth} p{0.65\textwidth}}
    \toprule
    \textbf{Training Set} & \textbf{Testing Set} \\
\bottomrule
    \end{tabular}
    \end{table}

\begin{table*}[htbp]
    \centering
    \caption{ACC on 59 new algorithms in Protocol-1.}
    \label{tab:acc-all}
    \footnotesize
    \renewcommand{\arraystretch}{1.0}
        \resizebox{1.0\textwidth}{!}{
        \begin{tabular}{l|cc|cc|cc|ccc|cc|ccc|c}
        \toprule
        \multirow{4}{*}{\textbf{Algorithm}} & \multicolumn{2}{c|}{\textbf{Basic Backbone}} & \multicolumn{2}{c|}{\textbf{Finetuned CLIP}} & \multicolumn{2}{c|}{\textbf{Reconstruction-based}} & \multicolumn{3}{c|}{\textbf{Frequency-domain}} & \multicolumn{2}{c|}{\textbf{Latent Space}} & \multicolumn{3}{c|}{\textbf{Representation}} & \multicolumn{1}{c}{\textbf{Aug Adv.}} \\
        \cmidrule{2-3} \cmidrule{4-16}
         & \textbf{Xception} & \textbf{Eff-B4} & \textbf{CLIP-L-FFT} & \textbf{Effort} & \textbf{RECCE} & \textbf{SBI} & \textbf{F3Net} & \textbf{SRM} & \textbf{SPSL} & \textbf{ProDet} & \textbf{LSDA} & \textbf{UCF} & \textbf{IID} & \textbf{CORE} & \textbf{sladd} \\
        & \cite{xception} & \cite{EfficientNet} & \cite{CLIP} & \cite{Effort} & \cite{recce} & \cite{SBI} & \cite{F3Net} & \cite{SRM} & \cite{SPSL} & \cite{ProDet} & \cite{LSDA} & \cite{UCF} & \cite{iid} & \cite{CORE} & \cite{sladd} \\
        \midrule
        ATVG-Net~\cite{ATVG-Net}         & 0.7110 & 0.7480 & 0.7601 & 0.8360 & 0.7412 & 0.8347 & 0.7588 & 0.7037 & 0.5620 & 0.6876 & 0.7553 & 0.7827 & 0.7287 & 0.7453 & 0.7494 \\
        FirstOrderMotion~\cite{FirstOrderMotion} & 0.7910 & 0.7955 & 0.8190 & 0.8679 & 0.7922 & 0.6973 & 0.8116 & 0.7681 & 0.6698 & 0.7347 & 0.8019 & 0.8256 & 0.7811 & 0.8069 & 0.7743 \\
        Talking-headVideo~\cite{TalkingHeadVideo} & 0.5541 & 0.5997 & 0.6177 & 0.7717 & 0.5593 & 0.5491 & 0.5514 & 0.5214 & 0.6152 & 0.6181 & 0.5996 & 0.5390 & 0.5203 & 0.5782 & 0.5863 \\
        MMreplacement~\cite{Forgerynet}    & 0.5252 & 0.5430 & 0.5837 & 0.5824 & 0.4954 & 0.4823 & 0.5072 & 0.5006 & 0.5359 & 0.4819 & 0.4863 & 0.5064 & 0.4751 & 0.5206 & 0.4821 \\
        DiscoFaceGAN~\cite{DiscoFaceGAN}     & 0.8069 & 0.8244 & 0.8138 & 0.8565 & 0.8161 & 0.7264 & 0.8394 & 0.7673 & 0.6821 & 0.7516 & 0.8086 & 0.8346 & 0.7829 & 0.8214 & 0.7948 \\
\bottomrule
        \end{tabular}
        }%
    \end{table*}

\begin{table*}[htbp]
    \centering
    \caption{AUC on 59 new algorithms in Protocol-1.}
    \label{tab:auc-all}
    \footnotesize
    \renewcommand{\arraystretch}{1.0}
        \resizebox{1.0\textwidth}{!}{
        \begin{tabular}{l|cc|cc|cc|ccc|cc|ccc|c}
        \toprule
        \multirow{4}{*}{\textbf{Algorithm}} & \multicolumn{2}{c|}{\textbf{Basic Backbone}} & \multicolumn{2}{c|}{\textbf{Finetuned CLIP}} & \multicolumn{2}{c|}{\textbf{Reconstruction-based}} & \multicolumn{3}{c|}{\textbf{Frequency-domain}} & \multicolumn{2}{c|}{\textbf{Latent Space}} & \multicolumn{3}{c|}{\textbf{Representation}} & \multicolumn{1}{c}{\textbf{Aug Adv.}} \\
        \cmidrule{2-3} \cmidrule{4-16}
         & \textbf{Xception} & \textbf{Eff-B4} & \textbf{CLIP-L-FFT} & \textbf{Effort} & \textbf{RECCE} & \textbf{SBI} & \textbf{F3Net} & \textbf{SRM} & \textbf{SPSL} & \textbf{ProDet} & \textbf{LSDA} & \textbf{UCF} & \textbf{IID} & \textbf{CORE} & \textbf{sladd} \\
        & \cite{xception} & \cite{EfficientNet} & \cite{CLIP} & \cite{Effort} & \cite{recce} & \cite{SBI} & \cite{F3Net} & \cite{SRM} & \cite{SPSL} & \cite{ProDet} & \cite{LSDA} & \cite{UCF} & \cite{iid} & \cite{CORE} & \cite{sladd} \\
        \midrule
        ATVG-Net~\cite{ATVG-Net}         & 0.8039 & 0.8392 & 0.9452 & 0.9808 & 0.8267 & 0.8875 & 0.8246 & 0.8346 & 0.7674 & 0.9519 & 0.8559 & 0.8725 & 0.8331 & 0.8392 & 0.8996 \\
        FirstOrderMotion~\cite{FirstOrderMotion} & 0.9016 & 0.8880 & 0.9734 & 0.9824 & 0.8925 & 0.7724 & 0.9000 & 0.8911 & 0.8559 & 0.8773 & 0.9034 & 0.9184 & 0.8844 & 0.9082 & 0.8910 \\
        Talking-headVideo~\cite{TalkingHeadVideo} & 0.5984 & 0.6486 & 0.6682 & 0.8397 & 0.5975 & 0.6163 & 0.6191 & 0.5438 & 0.6766 & 0.6627 & 0.6489 & 0.5970 & 0.5359 & 0.6316 & 0.6139 \\
        MMreplacement~\cite{Forgerynet}    & 0.5622 & 0.5732 & 0.6284 & 0.6336 & 0.5038 & 0.4873 & 0.5467 & 0.5026 & 0.5384 & 0.4746 & 0.4908 & 0.5493 & 0.4624 & 0.5562 & 0.4672 \\
        DiscoFaceGAN~\cite{DiscoFaceGAN}     & 0.9160 & 0.9263 & 0.9490 & 0.9605 & 0.9213 & 0.8242 & 0.9243 & 0.9143 & 0.8895 & 0.9117 & 0.9159 & 0.9217 & 0.8847 & 0.9199 & 0.9143 \\
\bottomrule
        \end{tabular}
        }%
    \end{table*}

\begin{table*}[h]
\centering
\renewcommand{\arraystretch}{1.2}
\resizebox{\linewidth}{!}{
\begin{tabular}{c p{14cm}}
\toprule
\textbf{Prompt ID} & \textbf{Textual Description Prompt} \\
\midrule
1 & The individual has shoulder-length brown hair styled with soft waves. They wear dangling earrings and a necklace, suggesting a formal or semi-formal occasion. Their expression is serious, with slightly parted lips as if mid-speech, indicating engagement in conversation. The posture appears upright, though only the head and shoulders are visible. The background is softly lit, featuring neutral tones that suggest an indoor setting, possibly a studio or event space. The overall appearance conveys professionalism and poise. \\
\midrule
2 & The individual has short, tousled brown hair and a stubbled beard, giving him a rugged appearance. His eyes are partially closed, and he has a slight smile, suggesting a relaxed or content mood. He is wearing a dark, casual shirt. The background is dimly lit, with indistinct shapes and colors, possibly indicating an evening event or gathering. His posture appears casual and confident. \\
\midrule
3 & The individual has short, graying hair with a receding hairline, a light beard, and a relaxed smile. His eyes are partially squinted, suggesting a friendly demeanor. He wears a black shirt and stands upright with a casual posture. The background features a blurred image of a desert landscape, possibly from a movie poster, indicating the setting might be a film-related event. The overall atmosphere is informal and approachable. \\
\midrule
4 & The individual has short, dark brown hair with a slight wave, styled neatly. Their face is oval-shaped with a strong jawline, and they have a calm, neutral expression. They are wearing a dark blue sweater over a white shirt. The posture is relaxed yet confident, leaning slightly forward. The background is blurred, suggesting an indoor setting with warm lighting, possibly a studio or home environment. The overall mood is serene and composed. \\
\bottomrule
\end{tabular}
}
\caption{Detailed textual description prompts (Prompt 1--10) adopted in our experiments.}
\label{tab:prompt_list}
\end{table*}



\end{document}